\documentclass[lettersize,journal]{IEEEtran}
\usepackage{amsmath,amsfonts}
\usepackage{algorithmic}
\usepackage{algorithm}
\usepackage{array}
\usepackage{subfigure}
\usepackage{textcomp}
\usepackage{stfloats}
\usepackage{url}
\usepackage{verbatim}
\usepackage{graphicx}
\usepackage{cite}
\usepackage{multirow}
\usepackage{makecell}
\usepackage{soul}
\usepackage[para,online,flushleft]{threeparttable}
\usepackage{booktabs}

\begin{document}

\title{Ground-VIO: Monocular Visual-Inertial Odometry with Online Calibration of Camera-Ground Geometric Parameters}

\author{Yuxuan Zhou, Xingxing Li, Shengyu Li, Xuanbin Wang, Zhiheng Shen
        % <-this % stops a space
%\thanks{This paper was produced by the IEEE Publication Technology Group. They are in Piscataway, NJ.}% <-this % stops a space
%\thanks{Manuscript received April 19, 2021; revised August 16, 2021.}
\thanks{This work was supported in part by the National Key Research and Development Program of China (2021YFB2501102), the National Natural Science Foundation of China (Grant 41974027), and the Sino-German mobility programme (Grant No. M-0054).\textit{(Corresponding author: Xingxing Li.)}}% <-this % stops a space
\thanks{The authors are with School of Geodesy and Geomatics, Wuhan University, China  (e-mail: xxli@sgg.whu.edu.cn).}
}

% The paper headers
\markboth{Manuscript}%
{Shell \MakeLowercase{\textit{et al.}}: A Sample Article Using IEEEtran.cls for IEEE Journals}

%\IEEEpubid{0000--0000/00\$00.00~\copyright~2021 IEEE}
%% Remember, if you use this you must call \IEEEpubidadjcol in the second
%% column for its text to clear the IEEEpubid mark.

\maketitle

\begin{abstract}
Monocular visual-inertial odometry (VIO) is a low-cost solution to provide high-accuracy, low-drifting pose estimation. However, it has been meeting challenges in vehicular scenarios due to limited dynamics and lack of stable features. In this paper, we propose Ground-VIO, which utilizes ground features and the specific camera-ground geometry to enhance monocular VIO performance in realistic road environments. In the method, the camera-ground geometry is modeled with vehicle-centered parameters and integrated into an optimization-based VIO framework. These parameters could be calibrated online and simultaneously improve the odometry accuracy by providing stable scale-awareness. Besides, a specially designed visual front-end is developed to stably extract and track ground features via the inverse perspective mapping (IPM) technique. Both simulation tests and real-world experiments are conducted to verify the effectiveness of the proposed method. The results show that our implementation could dramatically improve monocular VIO accuracy in vehicular scenarios, achieving comparable or even better performance than state-of-art stereo VIO solutions. The system could also be used for the auto-calibration of IPM which is widely used in vehicle perception. A toolkit for ground feature processing, together with the experimental datasets, would be made open-source\footnotemark[1].
\end{abstract}
\footnotetext[1]{https://github.com/GREAT-WHU/gv\_tools}

\section{Introduction}
\IEEEPARstart{V}{ision}-based solutions have been pivotal in the development of  intelligent vehicle applications \cite{ref_kitti}\cite{ref_autonomous_driving}. The low-cost camera could provide high-resolution texture information of the environment, enabling high-level  perception such as object detection \cite{ref_od} and scene parsing \cite{ref_sp}. On the other hand, visual simultaneous localization and mapping (VSLAM) provides a feasible approach for accurate vehicle pose estimation, which could be used for navigation tasks \cite{ref_vslam1}\cite{ref_vslam2}. Such aspect of vision-based navigation is later enhanced with the introduction of inertial measurement unit (IMU), bringing about better stability and accuracy with very limited additional expenses \cite{ref_vio}. Besides, IMU could resolve the scale ambiguity in monocular VSLAM, facilitating more practical use cases. The outstanding performance of visual-inertial odometry (VIO) and visual-inertial navigation system (VINS) has been demonstrated in unmanned aerial vehicle (UAV) applications \cite{ref_vinsmono}\cite{ref_smsckf}. 

However, as a possible low-cost navigation solution, monocular VIO has been meeting challenges when applied to ground vehicles. 
%One prerequisite for monocular VIO to maximize its estimation accuracy is sufficient excitation to the IMU.
Unlike UAVs, it is impractical for a ground vehicle to sufficiently excite the IMU during regular maneuvers. This would significantly affect the pose estimation performance of VIO due to lack of observability\cite{ref_observability1}\cite{ref_observability2}, especially for the scale which is unresolvable in monocular vision. Although stereo camera setups can mitigate this issue to some extent\cite{ref_observability3}, they entail additional expenses and computation cost. Other schemes turn to a higher integration level, introducing other sensors (e. g. wheel encoder, GNSS and LiDAR) to achieve better navigation performance\cite{ref_msf1,ref_msf2,ref_msf3}.

\begin{figure}[!t]
\centering
\includegraphics[width=8.4cm]{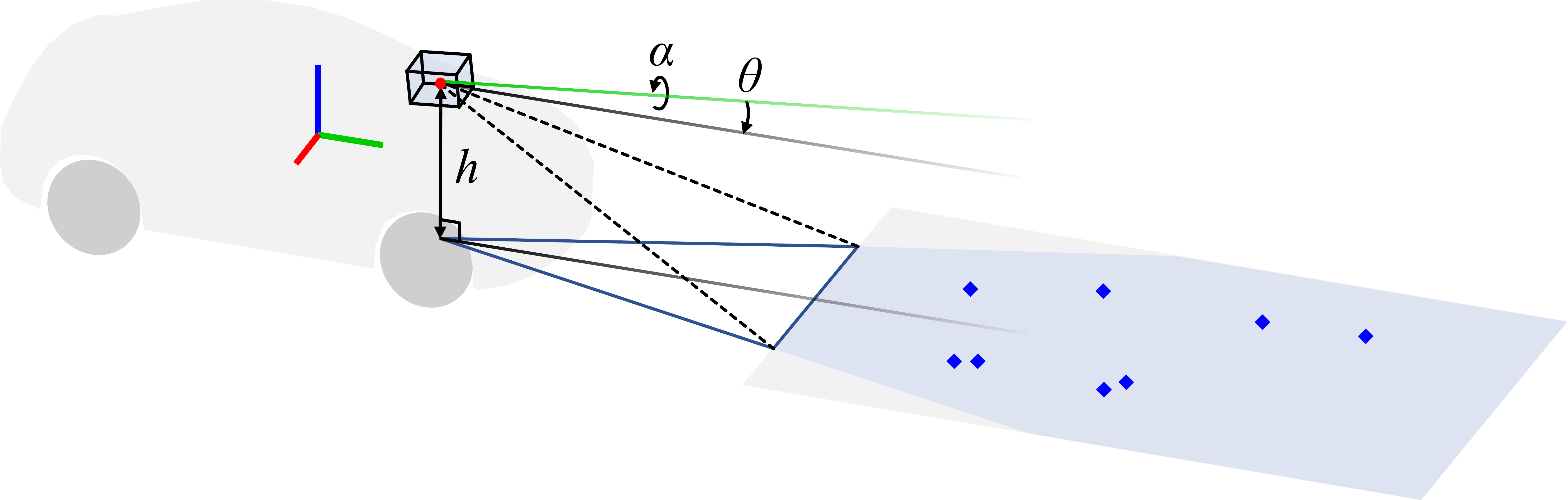}
\caption{Illustration of the camera-ground geometry. For a ground vehicle, the local ground plane could be expressed as a specific plane in the camera frame, which is parameterized as a height with a two-step rotation in this work (see Sect. III).}
\label{fig_cgg}
\end{figure}

It is noted that, the ground itself provides a natural and powerful constraint that could be utilized to enhance VSLAM/VIO performance. Considering the fact that the vehicle moves on the ground, the vehicle-mounted sensor and the local ground plane have a relatively fixed geometric relationship,  depending mainly on the sensor installation and the vehicle size. For VSLAM/VIO, the local ground plane could be expressed as a specific plane in the camera frame, as shown in Fig. \ref{fig_cgg}. We term this fixed relationship as the camera-ground geometry, which could be used to constrain the landmark depths, thu=s deriving metric-scale geometric information that is essential for high-accuracy pose estimation. However, few researches have given an in-depth insight into the application of camera-ground geometry in VIO.

In fact, such camera-ground geometry has been widely, sometimes implicitly, applied in vehicle perception. Typically, the well known inverse perspective mapping (IPM) technique\cite{ref_IPM1}\cite{ref_IPM2} is using the pre-calibrated camera-ground geometry to generate bird-eye view (BEV) images, or known as around-view monitoring (AVM)\cite{ref_avm}, thus to efficiently perceive the surrounding road environment\cite{ref_ipm_map1,ref_ipm_map2,ref_ipm_perception}. From this aspect, the online auto-calibration of the camera-ground geometry is also a meaningful issue and still remains unsolved.

In the proposed Ground-VIO, we introduce the online estimation of the camera-ground geometric parameters, termed as C-G parameters, into a monocular VIO, which could not only improve the odometry performance but also provide an approach for the auto-calibration of IPM. The contributions of this work are as follows:

\begin{itemize}
\item A vehicle-centered model is proposed to parameterize the camera-ground geometry, which is integrated into the monocular VIO for online calibration and to improve the navigation performance.
\item A novel visual front-end is developed to precisely track features on the ground by making use of the camera-ground geometry and IPM.
\item Both simulation tests and real-world experiments were conducted to validate different aspects of the system, including the estimation of C-G parameters, the odometry accuracy and the IPM calibration performance.
\item We make the ground feature processing module and the test data sequences open-source.
\end{itemize}

The rest of the paper is organized as follows. In Section II,
related works are discussed. In Section III, the system overview is presented. In Section IV, the core idea of camera-ground geometric model is explained. Section V presents the implementation details of Ground-VIO. The system performance is evaluated in Section VI and Section VII through simulation tests and real-world experiments respectively. The conclusion is finally given in Section VIII.

\section{Related Work}
\subsection{Visual-Inertial Odometry}
The aspect of VIO/VINS has been extensively investigated in the past decades, applied in both UAV and ground vehicle applications. Generally, the implementations could be divided into filter-based and optimization-based methods. For filter-based methods, the representative framework is multi-state constraint kalman filter (MSCKF)\cite{ref_msckf}, which maintains historical IMU poses in the state vector and uses common-view feature observations to construct geometric constraints among the poses. The variants of MSCKF have been developed to improve the framework by introducing observability constraint, extrinsic calibration\cite{ref_msckf2}, multi-IMU/camera configuration\cite{ref_mimc}, landmark states\cite{ref_openvins} and so on. For optimization-based implementations, the mainstream methods use a factor graph which jointly optimizes IMU preintegration factors \cite{ref_preintegration} and visual re-projection factors to estimate the navigation states and landmark positions\cite{ref_okvis}. These methods are expected to achieve better performance through iterations and relinearizations, in the expense of higher computation cost. The sliding window mechanism is a usual way to ensure real-time processing\cite{ref_vinsmono}\cite{ref_vinsfusion}, while some other implementations employ a local map to limit the problem size\cite{ref_orbslam3}.

Despite the advantages of low cost and high accuracy in ideal conditions, VIO/VINS has been meeting challenges when employed for vehicular applications due to limited dynamics, fast motion, and lack of stable features. To improve the practicality of such methods, it is necessary to introduce other sensors or fully utilize the inherent constraint of a ground vehicle.

\subsection{Vehicle Navigation Utilizing the Ground Constraint}
It is a natural idea to utilize the ground constraint when designing a navigation system for ground vehicles. Most of these researches model the ground as a local plane (or manifold) and constrain the vehicle motion on it, which is sometimes implicitly comprised in a vehicle-centered non-holonomic constraint (NHC). As pointed out in\cite{ref_se2_gf}, these implementations could be roughly divided into deterministic and stochastic SE(2) constraints. Deterministic SE(2) constraints strictly constrain the vehicle poses via parameterization or a deterministic model\cite{ref_lidar_se2}\cite{ref_nhc1}, while stochastic SE(2) constraints are applied to SE(3) pose estimation with a time-variant and probabilistic constraint\cite{ref_se2_1,ref_se2_2,ref_se2_3,ref_se2_4}. Generally, the former is more suitable for indoor or small-scale environments, while the latter shows superiority in outdoor environments with better resistance to outliers.  Although some of these methods use a vision-based sensor setup, most of them don't directly associate the visual observations with the ground. Differently, in\cite{ref_se2_gf}, stereo visual features are utilized to estimate the ground manifold representation, which is later used to constrain the vehicle pose.

Compared to the mentioned methods, our work focuses on the ground observed by the vehicle-mounted camera rather than the vehicle motion constrained on the ground. Actually, there is significant difference between ``the vehicle maneuvers on the ground'' and ``the observed features are on the ground''. For VSLAM/VIO, the latter statement could be used to constrain the landmark depths based on the relatively stable camera-ground geometry, thus to provide instantaneous scale-awareness to the system. This characteristic has been pointed out in \cite{ref_gc1,ref_gc2,ref_gc3}, all of which use the observed ground structure to realize scale-aware VSLAM based on a monocular camera. However, the camera height should be pre-calibrated in these methods, which limits the usability. Literature \cite{ref_gf} takes the camera-geometric geometry into the state vector of VIO, but discusses little about its mechanism.  In this work, we would demonstrate that the camera-ground geometry could be calibrated online in a monocular VIO without other infrastructure and could greatly improve the odometry performance.

\section{Camera-Ground Geometric Model}
In this section, the camera-ground geometric model utilized in the proposed system is firstly introduced.

For a camera mounted on a ground vehicle, it has a specific geometric relationship with the ground. Assuming the local ground is flat and the vehicle is a rigid body (temporarily ignoring the suspension system), the ground plane in the camera frame $c$ is a specific plane that could be unambiguously determined by its normal vector and distance\cite{ref_gc1}. 

\begin{figure}[!t]
\centering
\includegraphics[width=7.6cm]{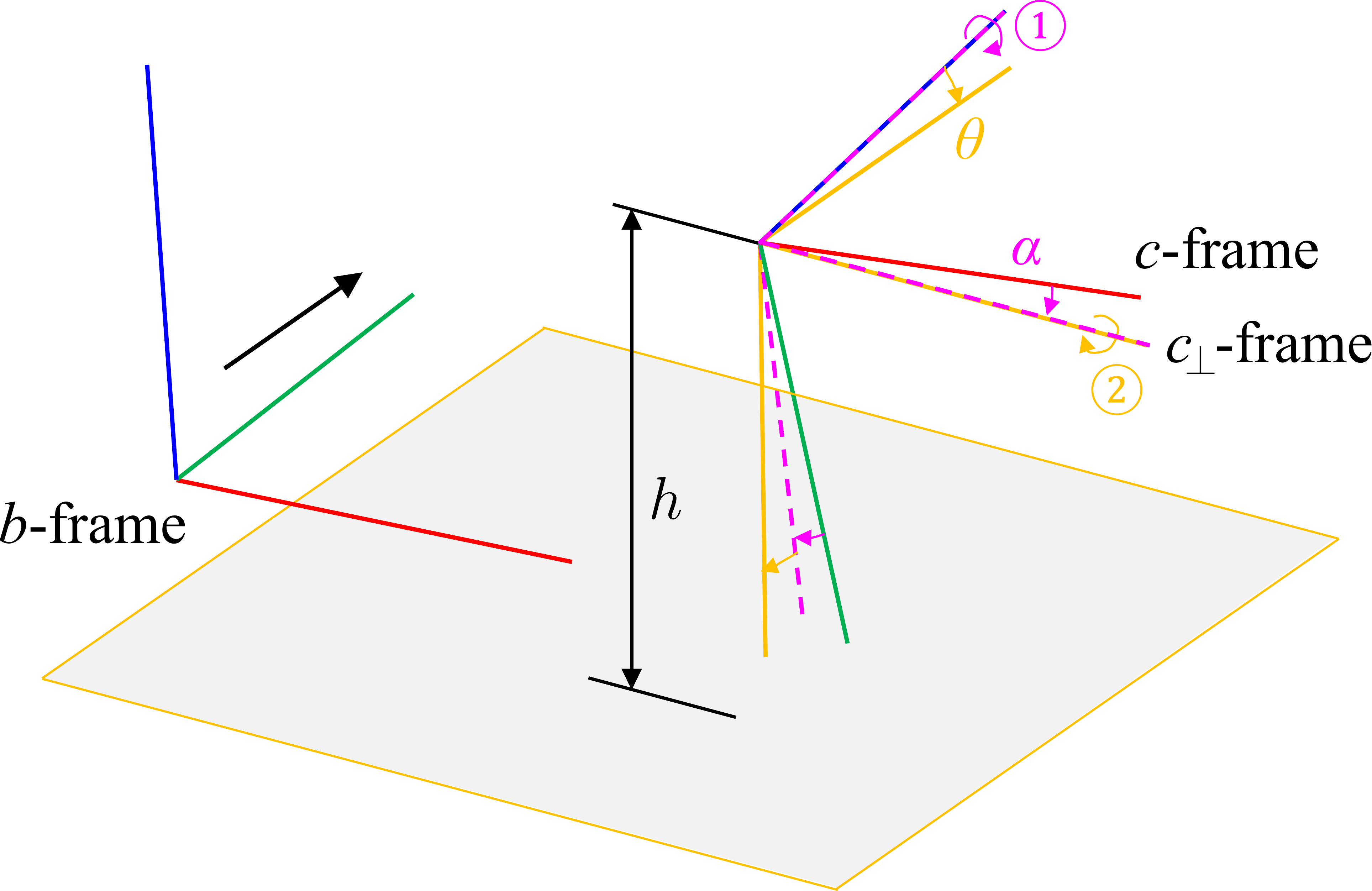}
\caption{Illustration of the C-G parameters. The $b$-frame is the vehicle body/IMU frame, $c$-frame is the camera frame, $c_\bot$-frame is the virtual camera frame whose X-Z plane is parallel with the local ground.}
\label{frame}
\end{figure}

Here, for better convenience, we parameterize the local ground plane using the height $h$ from the camera center to the ground and a two-step rotation which makes the X-Z plane of the camera frame parallel with the ground plane, as illustrated in Fig. \ref{frame}. Specifically, we firstly rotate the real camera frame $c$ around the Z-axis to make its X-axis parallel with the ground plane. Secondly, we rotate the obtained frame around the X-axis to get the expected virtual camera frame $c_{\bot}$, which could also be seen as the reference frame of IPM\cite{ref_ipm_map2}.

Thus, the ground plane in the camera frame could be expressed by the following one-row equation

\begin{equation}
\left({\mathbf{R}^{c}_{c_{\bot}}\left(\alpha,\theta\right)}^{\top}  \mathbf{p}^{c}_f\right)_{y} - h = 0
\label{equation_cg}
\end{equation}
with
\begin{align}
\setlength{\arraycolsep}{2.5pt}
\mathbf{R}^{c}_{c_{\bot}}\left(\alpha,\theta\right)=
\left[\begin{matrix}
\cos{\alpha} & -\sin{\alpha} & 0 \\
\sin{\alpha} & \cos{\alpha} & 0 \\
0 & 0 & 1
\end{matrix}\right]
\left[\begin{matrix}
1 & 0 & 0 \\
0 & \cos{\theta} & -\sin{\theta} \\
0 & \sin{\theta} & \cos{\theta}
\end{matrix}\right]
\end{align}
where $(\cdot)_{x/y/z}$ denotes the first/second/third row of a three-row vector/matrix, $\alpha$ and $\theta$ are the magnitudes of the two-step rotation, corresponding to the roll and the pitch angles of the camera. The triplet $\left(h,\theta,\alpha\right)$ is defined as the C-G parameters in this paper, which is similar to the parameterization in\cite{ref_gf}.

Such parameterization makes the estimation of camera-ground geometry straight-forward (i.e., one height and two angles), and it becomes easy to use IMU attitude to compensate the geometry (see Sect. IV-E). It is reasonable to expect the C-G parameters, which indicate the local ground plane, are statistically stable in common road environments without notable change of the sensor alignment or vehicle load. The proposed Ground-VIO fully utilizes this aspect, and several techniques would be given later to deal with complex conditions.

Given that a landmark $f$ in the environment is observed by the camera, we could get
\begin{align}
\mathbf{p}^{c}_f = \frac{\mathbf{u}_f}{\lambda_f}
,\ \ 
\mathbf{u}_f =\left[\begin{matrix}x\\y\\1\end{matrix}\right]={\pi}_c^{-1} \left[\begin{matrix}u\\v\\1\end{matrix}\right]
\label{equation_proj}
\end{align}
where $\left[\begin{matrix}u &v\end{matrix}\right]^{\top}$ is the pixel coordinates of $f$ on the image,  $\mathbf{u}_f = \left[\begin{matrix}x &y & 1\end{matrix}\right]^{\top}$ is the normalized image coordinates, ${\pi}_c$ is the camera projection matrix, ${\lambda}_f$ is the inverse depth of $f$.

Equation (\ref{equation_cg}) and (\ref{equation_proj})  reveal that, with known camera-ground geometry, the metric-scale (inverse) depth of a ground feature on the image could be instantaneously recovered
\begin{align}
\lambda_f = \frac{1}{h}  \left({\mathbf{R}^{c}_{c_{\bot}}\left(\alpha,\theta\right)}^{\top} \mathbf{u}_f\right)_y
\label{equation_lambda}
\end{align}
which yields great significance for monocular VSLAM/VIO.

\begin{figure}[!t]
\centering
\includegraphics[width=7.0cm]{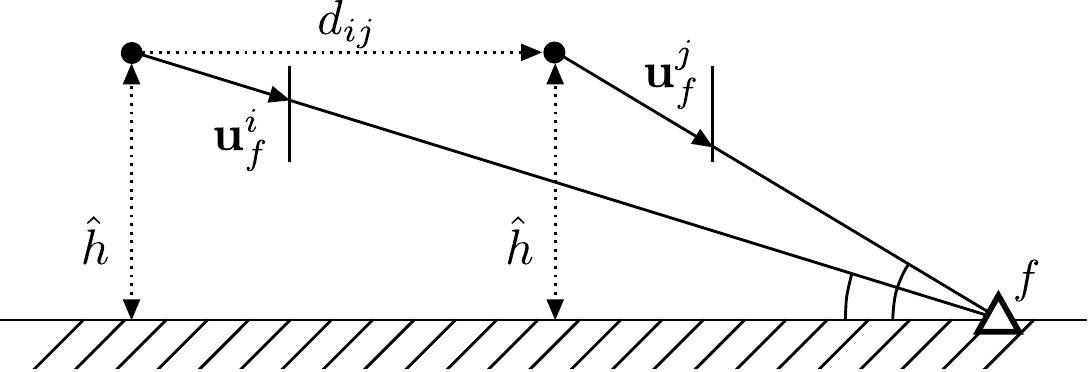}
\caption{A simple 1-D model to demonstrate how the camera-ground geometry works in VSLAM/VIO. }
\label{simple_model}
\end{figure}

By combining (\ref{equation_cg}) and (\ref{equation_proj}), the camera-ground geometry can be applied in VSLAM/VIO to constrain the landmark depths. Here, a simplified 1-D model is used to  qualitatively analyze how it makes sense in VSLAM/VIO. As shown in Fig. \ref{simple_model}, the odometry system needs to estimate the traveled distance $d_{ij}$ through the tracking of a landmark $f$. Assuming the C-G parameters are known and lead to a comprehensive camera height $\hat{h}$, the traveled distance $d_{ij}$ could be obtained 
\begin{align}
d_{ij} = \hat{h} \cdot \left({1}/(\mathbf{u}^i_f)_y-{1}/(\mathbf{u}^j_f)_y\right)
\end{align}
where $\mathbf{u}^i_f$ and $\mathbf{u}^j_f$ are observations of $f$ at epoch $i$ and $j$.
 
Assuming the visual observations are noiseless, the estimation of $d_{ij}$ only depends on the accuracy of $\hat{h}$
\begin{align}
d_{ij} \propto \hat{h}
\end{align}

Given a typical case of vehicular scenario where the comprehensive camera height $\hat{h}$ is 2 m with 2 cm error, the estimation of $d_{ij}$ would have 1\% relative error. In other words, the tracking of just one ground feature could derive geometric information about the translation with 1\% relative error, which is exceedingly meaningful for a monocular visual-inertial system.

It comes to the problems that, 1) how the camera-ground geometry could be integrated into the common VIO, and 2) how the C-G parameters could be obtained or estimated online. These problems would be explained in the following section.
\section{System Implementation}
In this section, the overview and implementation details of the proposed Ground-VIO will be presented.

\subsection{System Overview}

\begin{figure*}[!t]
\centering
\includegraphics[width=16.4cm]{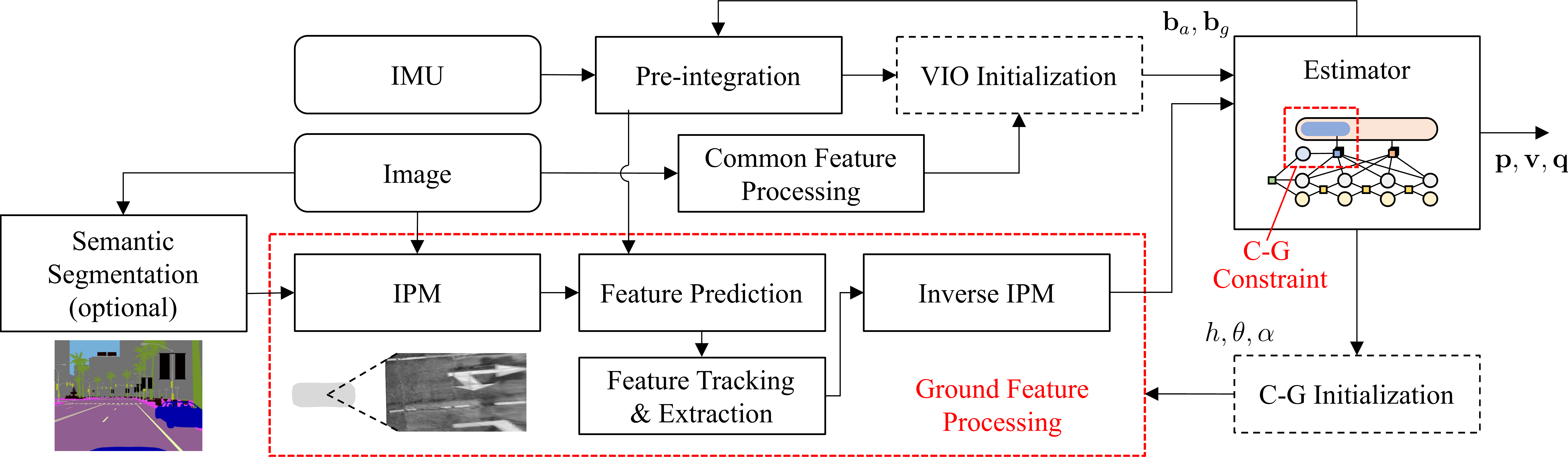}
\caption{Structure of the proposed system, adhering to the classic pipeline of optimization-based monocular VIO\cite{ref_vinsmono}. On this basis, a specially designed front-end for ground feature processing is added into the pipeline, which works in parallel with the common feature processor. The measurements from the IMU and the camera (including the ground features) are fused in the factor graph estimator. A module for the initialization of the C-G parameters is called periodically after the estimator until the parameters get initialized. }
\label{fig_system}
\end{figure*}

The overall structure of Ground-VIO is shown in Fig. \ref{fig_system}. The basic structure of the system follows the classic pipeline of optimization-based VIO \cite{ref_vinsmono} but with additional camera-ground-related mechanisms.

Basically, the collected images and IMU data are processed for common VIO initialization and optimization routines.

On this basis, an additional front-end is designed for ground feature processing and works in parallel with the common feature processor. The ground feature processor extracts and tracks features on the BEV images generated by IPM, which enables efficient and accurate tracking. A semantic segmentation module could be employed for ground segmentation but is not necessary, which would be discussed later.
%If a semantic segmentation module available, it could be used to determine the ground region on the image, but this is optional, which would be dicusssed later.
%An semantic segmentation module is optionally used to determine the region of ground feature extraction. Even without semantic segmentation, the incorrectly  extracted features would be mostly detected as outlier in later steps of ground feature tracking and factor graph optimization.

In factor graph optimization, the ground features are treated as a subset of visual features with additional camera-ground geometric constraints. These constraints could significantly improve the VIO performance and enable the online estimation of C-G parameters.

Under the situation that the C-G parameters are completely unknown at the beginning, the C-G initialization module would be called every time after common factor graph optimization. Once initialized, the camera-ground-related mechanisms in feature processing and factor graph optimization would be switched on. The C-G parameters would then be continuously refined during VIO running.

\subsection{Ground Feature Processing}
In typical implementations of VIO, feature points in the images are continuously tracked to construct visual measurements.
The proposed system follows the typical VIO routines\cite{ref_vinsmono} to detect and track environmental features in the camera view. To be specific, the feature detection method in\cite{ref_goodfeature} and KLT optical flow algorithm\cite{ref_klt} are employed, and a fundamental matrix-based RANSAC is used to detect outliers.

As to the ground features, their unique distribution and fast motions on the perspective image make them hard to track. The near-to-far ground plane is highly ``warped'' on the image, and the near points move drastically despite their better observation geometry.

Instead of the common method, we develop a special module for more precise data association of the ground features. 
It is noted that, with the camera-ground geometry, the 3-D position of every pixel on the perspective image corresponding to the ground could be instantaneously obtained, referring to (\ref{equation_lambda}). From another perspective, we could efficiently generate a BEV image using IPM, and every pixel on the image is directly related to a 3-D position. 
The following mapping relationship exists between the metric-scale 3-D point, the point on the perspective image and the point on the BEV image:
\begin{align}
\mathbf{p}^c_f=
\frac{1}{\lambda}_f\cdot\left[\begin{matrix}
x \\
y \\
1
\end{matrix}\right]   = h\cdot {\mathbf{R}^{c}_{c_{\bot}}\left(\alpha,\theta\right)}
\left[\begin{matrix}
x_{\bot} \\
1 \\
-y_{\bot}
\end{matrix}\right]
\label{equation_mapping}
\end{align}
where  $\mathbf{u}_{\bot}=\left[\begin{matrix}x_{\bot} & y_{\bot} & 1\end{matrix}\right]^{\top}$ is the normalized image coordinates of $f$ on the BEV image, the inverse depth $\lambda_f$ refers to (\ref{equation_lambda}). The generation of BEV images through IPM refers to\cite{ref_ipm_map2}.

%Then, the feature extraction/tracking could be performed on the BEV image.
The knowledge of camera-ground geometry makes the accurate prediction of ground feature tracking possible. Every time a new image comes, we could predict the position of an existing ground feature with the help of the IMU-predicted relative pose
\begin{align}
\mathbf{p}^{c_{k+1}}_f = \hat{\mathbf{R}}^{c_{k+1}}_{c_k} \mathbf{p}^{c_k}_f + \hat{\mathbf{p}} ^{c_{k+1}}_{c_k}
\label{equation_pred}
\end{align}
where $\left(\hat{\mathbf{R}}^{c_{k+1}}_{c_k}, \hat{\mathbf{p}} ^{c_{k+1}}_{c_k}\right)$ is the relative pose estimated by IMU integration. Combining (\ref{equation_mapping}) with (\ref{equation_pred}), the prediction of ground features could be performed  on either the perspective image or the BEV image, which could limit the search region of optical flow tracking to several pixels, thereby greatly improving the tracking performance. 

\begin{figure}[!t]
\centering
\includegraphics[width=7.7cm]{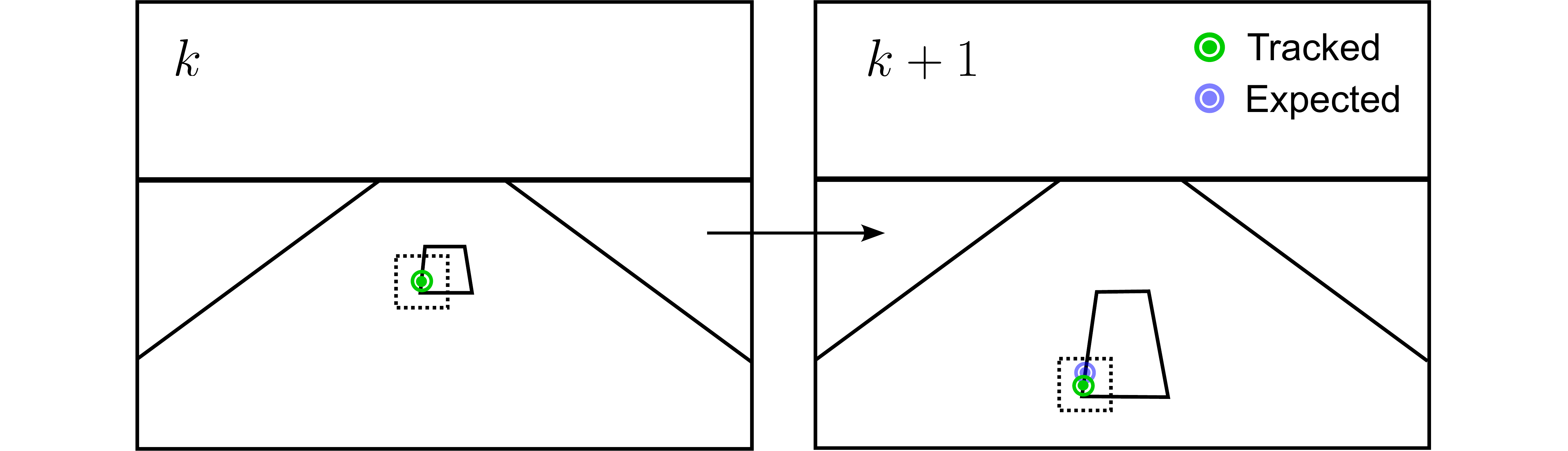}
\caption{Illustration of a failure case of optical flow tracking on the perspective image. The green scatter denotes the tracked feature, while the blue scatter is the expected accurate correspondence.}
\label{klt_failure}
\end{figure}

In Ground-VIO, we choose to extract and track features on the BEV image, for the reason that the BEV image is less ``warped'' and has better tracking consistency. In fact, the KLT optical flow tracking doesn't guarantee scale and rotation invariance, with a failure case illustrated in Fig. \ref{klt_failure}. Fortunately, the IPM could recover the metric-scale geometry of ground features and eliminate most of the scaling effect during fast motion, thus contributing to better tracking precision. Fig. 6 illustrates the tracking of ground features on the BEV image with IMU-aided feature prediction. In addition, a homography matrix-based RANSAC method is used to efficiently detect outlier feature trackings\cite{ref_cv_calib}. In our implementation, we mainly focus on the rectangle area (15 m far and ${\pm}$ 3 m wide with 0.015 m spatial resolution) in front of the vehicle-mounted camera, which facilitates multi-frame tracking of the ground features during regular vehicle maneuvers and guarantees good tracking precision.

\begin{figure}[!t]
\centering
\includegraphics[width=8.2cm]{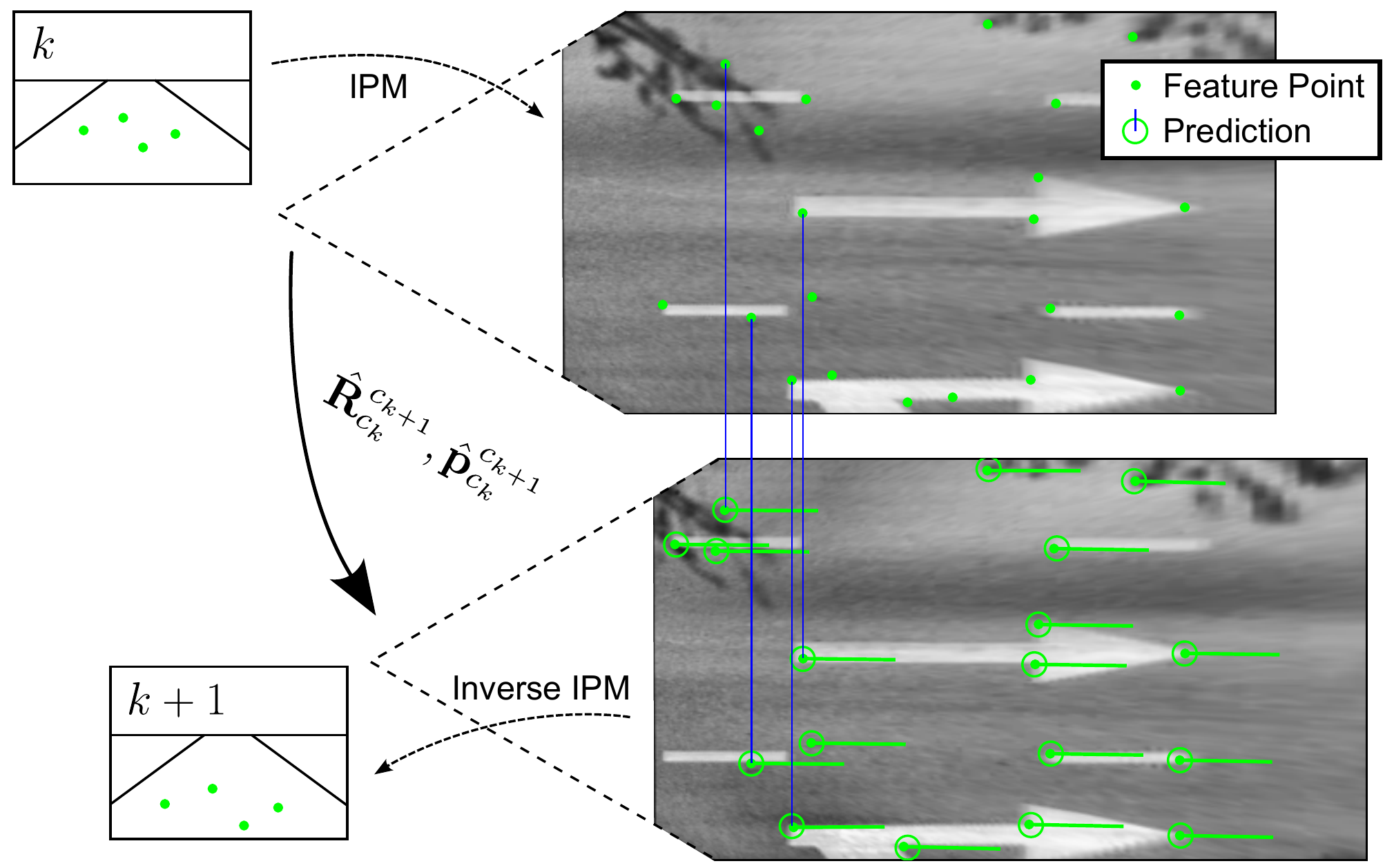}
\caption{Ground feature tracking on the BEV image. }
\label{fig_bev_tracking}
\end{figure}

After the feature processing on the BEV image, the obtained ground features are re-mapped to the perspective image through the inverse process of IPM, thus these features could be processed consistently with the common features. By doing so, the tracking on the BEV image helps improve the tracking precision without introducing systematic errors related to the C-G parameters. During the operation of VIO, the C-G parameters used for ground feature processing could be continuously updated.

Intuitively, to extract and track the ground features, it is needed to specifically identify the ground region in the image. This is not a hard task by applying deep learning-basd semantic segmentation\cite{ref_ss1}\cite{ref_ss2}, which performs well in vehicular scenarios. Yet in the proposed method, the semantic segmentation is optional.
The IPM processing itself could exclude most objects that are not on the ground surface, and the accurate feature prediction based on C-G parameters plus the RANSAC method could exclude outliers on the BEV image (e. g. vehicles, guardrails). Later in factor graph optimization, the influence of gross errors could be further mitigated through outlier detection methods. Therefore, although semantic segmentation could contribute to best performance of the system, it is not necessary.

In factor graph optimization, the extracted ground features are treated as a subset of visual features to construct the visual re-projection factors, while additional camera-ground constraints would be applied to them.

\subsection{Optimization-based Visual-Inertial Odometry}
Adhering to\cite{ref_vinsmono}, we maintain a sliding-window factor graph to simultaneously estimate the navigation states, landmarks and, additionally, the C-G parameters by optimizing different kinds of measurements. 

The state vector of Ground-VIO is defined as follows
\begin{align}
\mathcal{X}&= \left(\mathbf{x}_0 ,\ \mathbf{x}_1 ,\ \cdots ,\ \mathbf{x}_n ,\ \mathbf{x}_{\bot} ,\ {\lambda}_0 ,\ {\lambda}_1 ,\ \cdots ,\ {\lambda}_m \right) \\
\mathbf{x}_k&= \left(\mathbf{p}^w_{b_k} ,\ \mathbf{q}^w_{b_k} ,\ \mathbf{v}^w_{b_k} ,\ \mathbf{b}_{a,b_k},\ \mathbf{b}_{g,b_k}\right),\ k \in \left[0,\ n\right] \\
\mathbf{x}_{\bot} &= \left( h,\ \theta,\ \alpha\right)
\end{align}
where  $\mathbf{p}^w_{b_k}$, $\mathbf{q}^w_{b_k}$, $\mathbf{v}^w_{b_k}$ are the position, attitude and velocity of the $k$-th frame expressed in the world frame, $\mathbf{b}_{a,b_k}$ and $\mathbf{b}_{g,b_k}$ are the accelerometer bias vector and the gyroscope drift vector, ${\lambda}_0$, ${\lambda}_1$, $\cdots$, ${\lambda}_m$ are the inverse depths of the landmarks. Each landmark is anchored in the first observation frame within the sliding window.

\begin{figure}[!t]
\centering
\includegraphics[width=8.4cm]{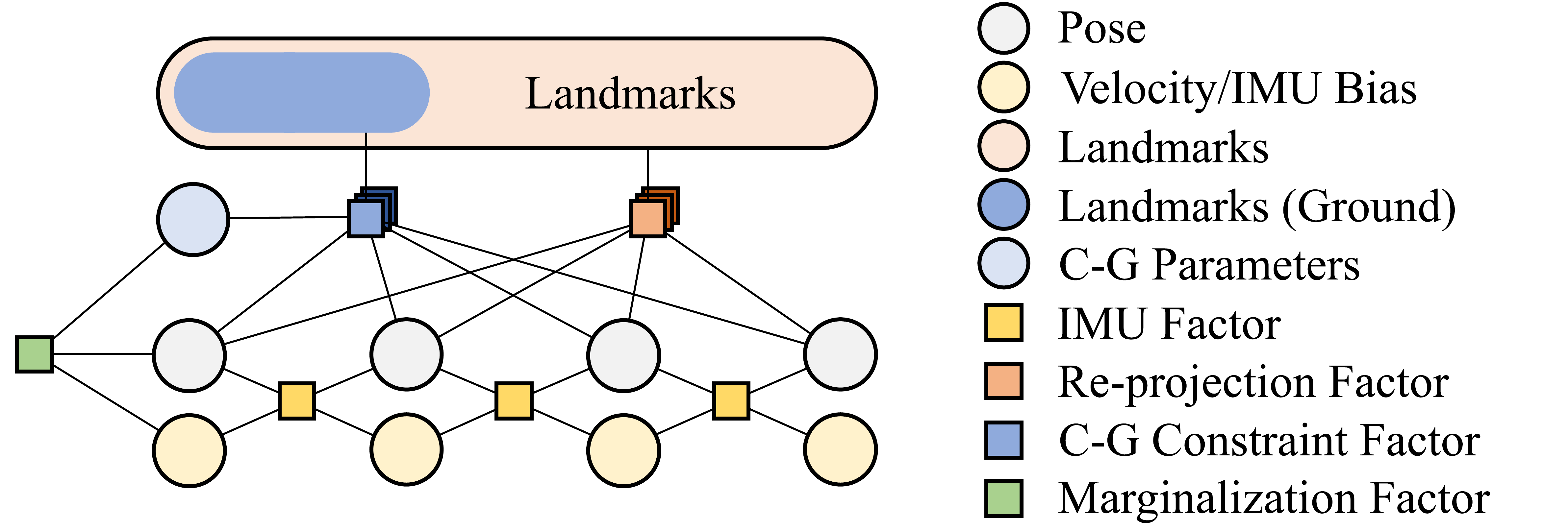}
\caption{The factor graph optimization of Ground-VIO. }
\label{fig_1}
\end{figure}

The following factors are considered in the optimization:

\textbf{1) IMU preintegration factor:}

The IMU data between frames are preintegrated and used to construct the IMU preintegration factors.
The residual could be expressed as
\begin{align}
\begin{split}
&\mathbf{r}_{\text{IMU}}\left(\hat{\mathbf{\alpha}} ^{b_k}_{b_{k+1}},\hat{\mathbf{\beta}} ^{b_k}_{b_{k+1}},\hat{\mathbf{\gamma}} ^{b_k}_{b_{k+1}},\mathbf{x}_k,\mathbf{x}_{k+1}\right)= \\
&\left[\begin{matrix}
{\mathbf{R}^w_{b_k}}^{\top} \left( \mathbf{p}^w_{b_{k+1}} - \mathbf{p}^w_{b_k}+ \frac{1}{2}\mathbf{g}^w\Delta{t^2_k} - \mathbf{v}^w_{b_k} \Delta t_k - \hat{\mathbf{\alpha}}^{b_k}_{b_{k+1}} \right) \\
{\mathbf{R}^w_{b_k}}^{\top} \left( \mathbf{v}^w_{b_{k+1}} + \mathbf{g}^w \Delta t_k - \mathbf{v}^w_{b_k}\right) - \hat{\mathbf{\beta}} ^{b_k}_{b_{k+1}}  \\
2\left[ {\mathbf{q}^w_{b_k}}^{-1} \otimes \mathbf{q}^w_{b_{k+1}} \otimes {\left(\hat{\mathbf{\gamma}} ^{b_k}_{b_{k+1}}\right)^{-1}} \right]_{xyz} \\
\mathbf{b}_{a,b_{k+1}} - \mathbf{b}_{a,b_{k}} \\
\mathbf{b}_{g,b_{k+1}} - \mathbf{b}_{g,b_{k}}
\end{matrix}\right]
\end{split}
\end{align}
where $k$ and $k+1$ are the epochs of adjacent frames, $\Delta t_k$ is the time interval, $\hat{\mathbf{\alpha}} ^{b_k}_{b_{k+1}}$, $\hat{\mathbf{\beta}} ^{b_k}_{b_{k+1}}$, $\hat{\mathbf{\gamma}} ^{b_k}_{b_{k+1}}$ are the IMU preintegration terms\cite{ref_vinsmono}. 

The IMU measurements provide stable relative pose information based on the navigation state estimation, but they alone couldn't measure the absolute values of the translation and the velocity. When combined with visual measurements in VIO, metric-scale translation/velocity could be derived as long as the IMU is sufficiently excited\cite{ref_observability1}. 

\textbf{2) Visual re-projection factor:}

The visual features maintained in the sliding window, including the ground features, are used to construct the visual re-projection factors. The residual could be expressed as
\begin{align}
\mathbf{r}_{\text{cam}}\left( \mathbf{u}^i_f,\mathbf{u}^j_f, \mathbf{x}_i, \mathbf{x}_j, {\lambda}_f \right)=
\left[ 
\begin{matrix}
(\mathbf{p}^{c_j}_f)_x/(\mathbf{p}^{c_j}_f)_z \\
(\mathbf{p}^{c_j}_f)_y/(\mathbf{p}^{c_j}_f)_z
\end{matrix}
\right]
- \mathbf{u}^j_f
\end{align}
with
\begin{align}
\mathbf{p}^{c_j}_f ={\mathbf{R}^w_{b_j} }^{\top}\left( {\mathbf{R}^w_{b_i}}\left(\mathbf{R}^b_c\left(\frac{ \mathbf{u}^i_f}{{\lambda}_f}\right) + \mathbf{p}^b_c\right) + \mathbf{p}^w_{b_i} - \mathbf{p}^w_{b_j}\right)
\label{equation_anchor2target}
\end{align}
where $\mathbf{u}^i_f$ and $\mathbf{u}^j_f$ are visual observations of $f$ at epoch $i$ and $j$, $\left(\mathbf{R}^b_c,\ \mathbf{p}^b_c\right)$ are the IMU-camera extrinsic parameters. 

The visual measurements are used to strongly constrain the vehicle poses and landmark positions through a bundle-adjustment (BA)-like model.

\textbf{3) Camera-ground constraint factor:}

\begin{figure}[!t]
\centering
\includegraphics[width=8.4cm]{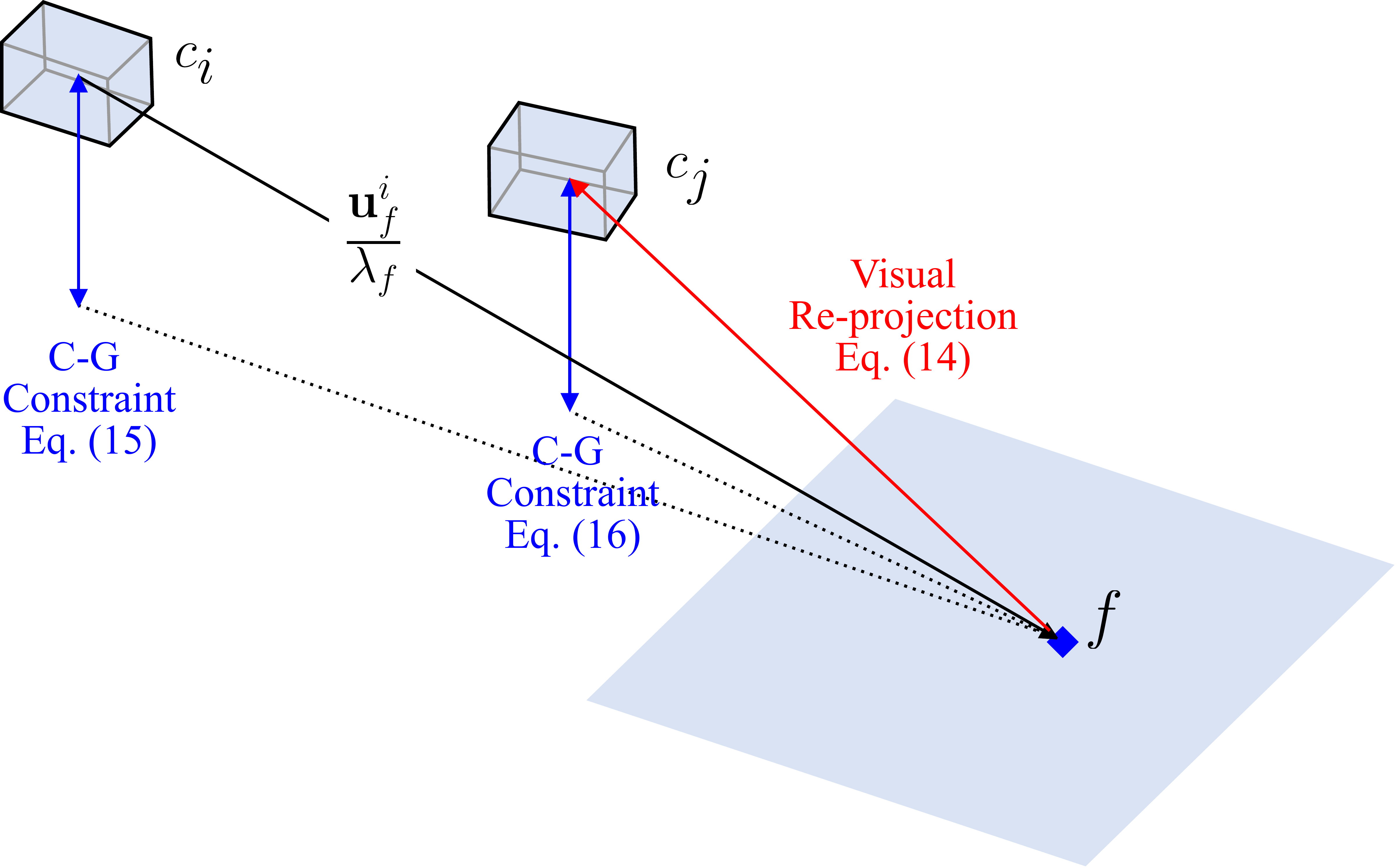}
\caption{Illustration of the visual re-projection factor and camera-ground constraint factors constructed upon a ground landmark $f$. Here, $c_i$ is the anchor frame, while $c_j$ is the target frame.}
\label{fig_bevxxxx}
\end{figure}

Camera-ground constraints are applied to the ground features maintained in the sliding window, based on the model in Sect. III.
In our implementation, there are two kinds of camera-ground constraint factors, depending on the anchor frame of the ground feature and the frame that the camera-ground constraint is applied, termed as the target frame. 

If the anchor frame and the target frame are the same, the residual could be expressed as
\begin{align}
r_{\text{C-G}}\left(\mathbf{u}^i_f,\lambda_f,\mathbf{x}_{\bot}\right)= h -\left( {\mathbf{R}^{c}_{c_{\bot}}\left(\alpha,\theta\right)}^\top \frac{\mathbf{u}^i_f}{\lambda_f}\right)_y 
\label{equation_single_frame}
\end{align}

If the anchor frame and the target frame are different, the residual could be expressed as
\begin{align}
\begin{split}
r_{\text{C-G}}\left(\mathbf{u}^i_f,\lambda_f,\mathbf{x}_i,\mathbf{x}_j,\mathbf{x}_{\bot}\right)= h -\left({\mathbf{R}^{c}_{c_{\bot}}\left(\alpha,\theta\right)}^\top\mathbf{p}^{c_j}_f\right)_y
\end{split}
\end{align}
where the $i$-th frame is the anchor frame, the $j$-th frame is the target frame, and $\mathbf{p}^{c_j}_f$ refers to (\ref{equation_anchor2target}).

By introducing the camera-ground geometric constraints into the estimator, the C-G parameters could be estimated and refined based on the information gained by VIO. Once the C-G parameters get converged, the constraints could reciprocally provide driftless, metric-scale geometric information to VIO.
The mechanism of ``make some parameters converge and use it to maintain the estimation performance'' is similar to the IMU biases.
Yet the converged C-G parameters are expected to have a more sustained influence, for: 1) the IMU biases are time-variant but the C-G parameters are relatively stable, 2) the C-G parameters are at the same order with pose estimation which don't need integration like IMU.

The optimization problem could then be expressed as minimizing above residuals and prior terms following
\begin{align}
\begin{split}
\mathop{\min}\limits_{\mathcal{X}}
\Big\{
&\left\| \mathbf{r}_p - \mathbf{H}_p \mathcal{X}
\right\|^2 +\\
\sum_{k\in \left[0,n\right)}&{
\left\|\mathbf{r}_{\text{IMU}}\left(\hat{\mathbf{\alpha}} ^{b_k}_{b_{k+1}},\hat{\mathbf{\beta}} ^{b_k}_{b_{k+1}},\hat{\mathbf{\gamma}} ^{b_k}_{b_{k+1}},\mathbf{x}_k,\mathbf{x}_{k+1}\right)\right\|^2_{\mathbf{P}_\text{IMU}}} +\\
\sum_{i< j\in \left[0,n\right],f \in \mathcal{F}}&{
\rho_\text{H}
(\left\|
\mathbf{r}_{\text{cam}}\left( \mathbf{u}^i_f,\mathbf{u}^j_f, \mathbf{x}_i, \mathbf{x}_j, {\lambda}_f 
\right)
\right\|^2_{\mathbf{P}_\text{cam}})} +\\
\sum_{i\in \left[0,n\right],f \in \mathcal{F}_{\bot}}&{
\rho_\text{C}
(\left\|
r_{\text{C-G}}\left(\mathbf{u}^i_f,\lambda_f,\mathbf{x}_{\bot}
\right)
\right\|^2_{P_\text{C-G}})} +\\
\sum_{i< j\in \left[0,n\right],f \in \mathcal{F}_{\bot}}&{
\rho_\text{C}
(\left\|
r_{\text{C-G}}\left(\mathbf{u}^i_f,\lambda_f,\mathbf{x}_i,\mathbf{x}_j,\mathbf{x}_{\bot}
\right)
\right\|^2_{P_\text{C-G}})}\Big\}
\end{split}
\end{align}
where $(\mathbf{r}_p, \mathbf{H}_p)$ is the prior information obtained from marginalization\cite{ref_vinsmono},
$\mathcal{F}$ is the set of landmarks maintained in the sliding window, $\mathcal{F}_{\bot}$ is the set of ground landmarks, which is a subset of $\mathcal{F}$,
$\rho_\text{H}(\cdot)$ and $\rho_\text{C}(\cdot)$ are Huber and Cauchy kernel functions\cite{ref_ceres} respectively, 
$\mathbf{P}_\text{IMU}$, $\mathbf{P}_\text{cam}$, $P_\text{C-G}$ are covariances/variances of the residuals. The ceres-solver\cite{ref_ceres} is employed to solve the optimization problem.

\subsection{Initialization of Camera-Ground Parameters}
If the C-G parameters are completely unknown at the beginning, the ground feature processing module couldn't work properly and it is hard to construct accurate camera-ground constraint factors. In this case, the system needs to online initialize the C-G parameters.

It is recognized that the monocular VIO has the capability to perceive metric-scale environmental structure with enough IMU excitation\cite{ref_observability1}. On this basis, it is completely possible to online initialize the C-G parameters without auxiliary information from other sensors. The specific procedure of the initialization is presented as follows.

After the VIO initialization, the common VIO routines start to work. So far, without the knowledge of C-G parameters, the ground features could only be tracked on the perspective image (without IPM processing). To achieve this, a conservative region of interest (ROI) on the image is used, which is determined by the IMU-camera extrinsics and a rough vehicle height. For the initialization of C-G parameters, the uncertainties (i. e., variances) of the ground landmarks in the sliding window are periodically checked. Once enough ground landmarks below the uncertainty threshold are obtained, observations of these landmarks are stacked together to estimate the C-G parameters, following
\begin{align}
\begin{split}
&\left(\hat{h},\hat{\theta},\hat{\alpha}\right) =
\mathop{\arg\min}\limits_{\left(h,\theta,\alpha\right)}\\
&\sum_{
i\le j\in\left[0,n\right],f\in {F}_\bot
}{
\left\| h - 
\left[
{\mathbf{R}^{c}_{c_{\bot}}\left(\alpha,\theta\right)}^{\top}\left(\mathbf{R}^{c_j}_{c_i}\frac{\mathbf{u}^{c_i}_f }{\lambda_f}+ \mathbf{p}^{c_j}_{c_i}\right)
\right]_y
\right\|
}
\end{split}
\end{align}

After the initialization of C-G parameters, the ground feature processing module would be switched on for better tracking accuracy. At the same time, the camera-ground constraint factors would be applied in the factor graph to enhance VIO, and the C-G parameters will be further refined.

\subsection{Dealing with Complex Road Conditions}
\begin{figure}[t]
\centering
\subfigure[]{
\includegraphics[width=3.9cm]{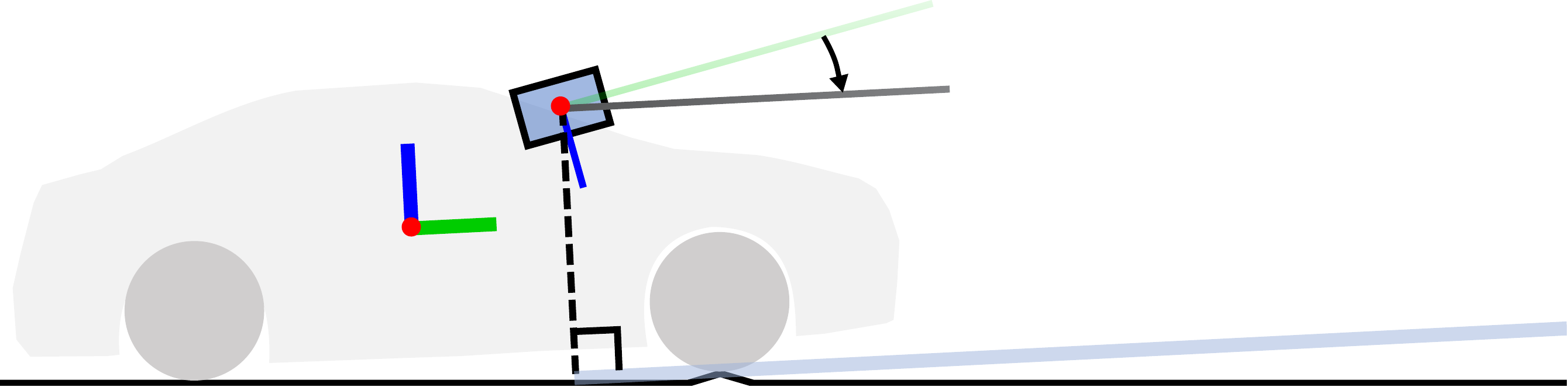}
}
\quad
\subfigure[]{
\includegraphics[width=3.9cm]{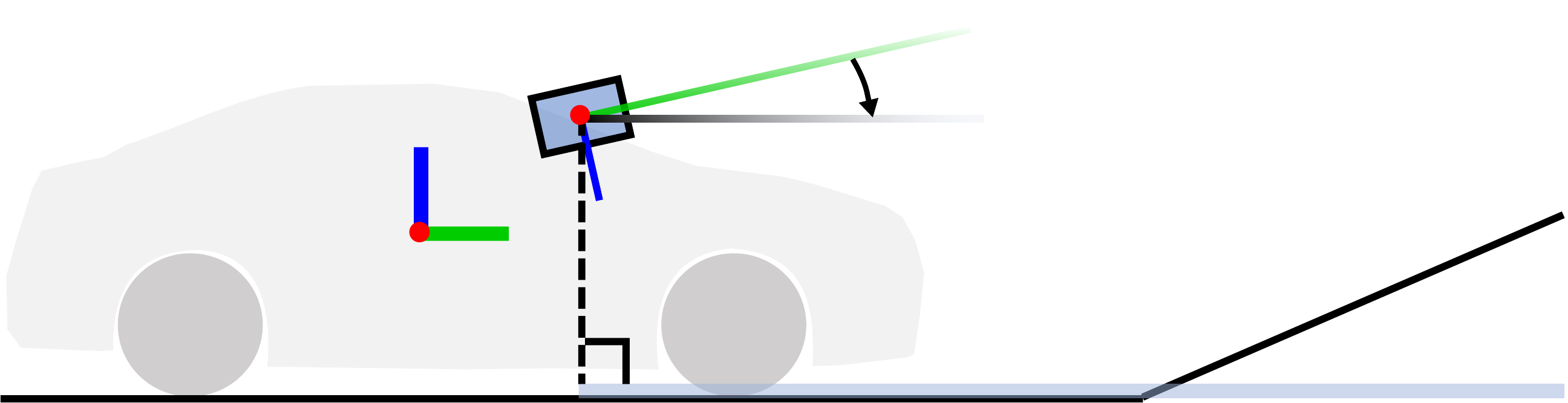}
}
\caption{Two typical cases of complex road conditions, (a) attitude vibration caused by road irregularity and vehicle dynamics. (b)  change of the road slope.}
\label{complex_condition}
\end{figure}

In real-world scenarios, the road conditions could be relatively complex and don't conform to the ideal camera-ground geometric model depicted in Sect. III. Such conditions could be mainly covered by the following two cases: 1) attitude vibration of the vehicle caused by road irregularity or vehicle dynamics, 2) change of the road slope. These two cases are illustrated in Fig. \ref{complex_condition}, which would lead to systematic errors and affect the system performance if not carefully considered.

\begin{figure}[!t]
\centering
\includegraphics[width=8.4cm]{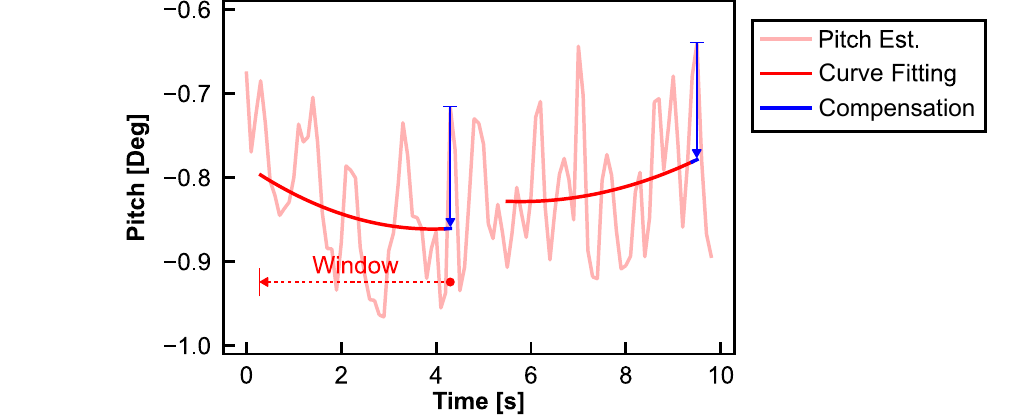}
\caption{An example of IMU-based pitch compensation when applying the camera-ground geometric constraint in realistic scenarios.}
\label{pitch_comp}
\end{figure}

In the proposed system, several tricks are employed to mitigate the effect of these problems. To deal with high-frequency attitude vibration of the vehicle (Fig. \ref{complex_condition}(a)), we use the local IMU attitude estimation to compensate the C-G parameters temporarily, as shown in Fig. \ref{pitch_comp}.
In our implementation, only the pitch component $\theta$ is compensated, which is more sensitive considering the ground region of interest (${\pm}$ 3 m wide, 15 m far).
To be specific, we use a 4-second window of historical pitch estimation to fit a quadratic curve and to calculate the pitch compensation of the current epoch, following
\begin{align}
{\theta}_{comp} = {\theta}^w_{b_k} - \hat{\theta}^w_{b_k}
\end{align}
where ${\theta}^w_{b_k}$ is the current IMU pitch estimation, $\hat{\theta}^w_{b_k}$ is the pitch predicted by curve fitting.
And when applying the camera-ground constraint factors in this frame, the C-G parameter $\theta$ is compensated temporarily
\begin{align}
{\theta}_k = {\theta} + {\theta}_{comp}
\end{align}
where ${\theta}_k$ is the taken as the temporary C-G parameter at epoch $k$. By doing so, the noise of the camera-ground constraint caused by attitude vibration could be significantly mitigated.

To deal with the change of the road slope (Fig. \ref{complex_condition}(b)), firstly the ground feature processing could abandon some of the feature observations that don't conform to the planar ground assumption. Secondly, when constructing the camera-ground factors, we use a relatively strict outlier removal strategy, with a cut-off threshold  plus a Cauchy kernel function, in order to counter gross errors caused by drastic slope changes.

\section{Simulation Tests}
Simulation tests are conducted to evaluate the system performance in relatively ideal conditions. The advantage of simulation is that the vehicle-sensor alignment and the environmental geometry are precisely known, which facilitates more in-depth analysis.
The CARLA simulator\cite{ref_carla}, which provides exquisite 3-D scenes and realistic vehicle dynamics, is used to generate the vehicle poses and  images. The IMU data is separately simulated based on B-spline fitting\cite{ref_openvins} of the 10 Hz ground truth poses, where custom biases and noises are added. The settings of the simulation are listed in TABLE \ref{table_simu_setting}.

\begin{table}[h]
\caption{Sensor Settings of the Simulation Tests.}
\label{table_simu_setting}
\begin{center}
\begin{threeparttable}
\begin{tabular}{l|l}
\hline
Simulation settings&   \\
\hline
Image resolution&\makecell[c]{$1024 \times 768$}\\
Field of view (FOV)&  \makecell[c]{$60{^\circ}$}\\
Image frequency &  \makecell[c]{$10$}\\
IMU frequency &  \makecell[c]{$100$}\\
Velocity random walk (VRW) &  \makecell[c]{${0.12\ (m/s/\sqrt{hr})}$}\\
Angle random walk (ARW) &  \makecell[c]{${0.5\ (^\circ/\sqrt{hr})}$}\\
Accelerometer bias\textsuperscript{1} &  \makecell[c]{$\left(1000,1000,1000\right)\ (mGal)$}\\
Gyroscope drift\textsuperscript{1} &  \makecell[c]{$\left(100,100,100\right)\ (^\circ/hr)$}\\
\hline
\end{tabular}
\begin{tablenotes}
\item[1] For simplicity, only constant biases are considered.
\end{tablenotes}
\end{threeparttable}
\end{center}
\end{table}
\begin{figure}[!t]
\centering
\includegraphics[width=7.4cm]{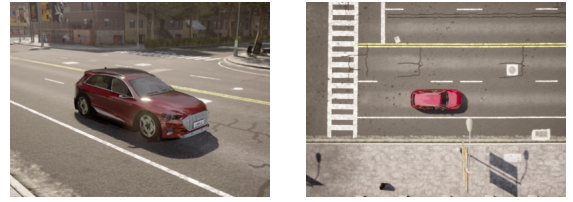}
\caption{The vehicle actor and the 3-D environment applied in the CARLA simulator. }
\label{fig_1}
\end{figure}
\begin{figure}[!t]
\centering
\includegraphics[width=8.4cm]{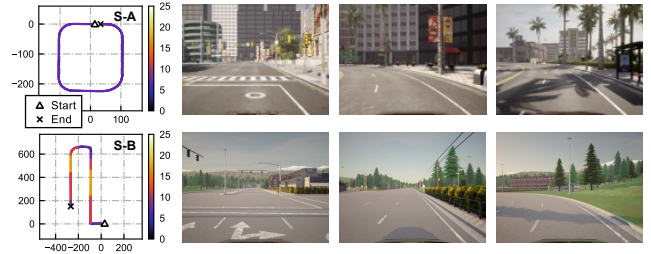}
\caption{Simulation trajectories and example images. The colorbar indicates the vehicle speed (m/s). Top: Seq. S-A, which uses the ``Town10'' world in CARLA simulator. The road texture is distinct and environmental features are rich in this world. Bottom: Seq. S-B, which uses the ``Town06'' world in CARLA simulator. Fewer distinct ground features, mainly the road markings, could be observed in this world.}
\label{simu}
\end{figure}

The vehicle trajectories and the captured images in the simulation tests are shown in Fig. \ref{simu}. The simulation consists of two sequences, namely S-A and S-B, corresponding to urban and highway environments respectively. The vehicle dynamics caused by the suspension system are considered, leading to up to ${\pm}$ 0.5$^\circ$ vibration of the vehicle attitude. 

Different schemes of VIO are tested on the simulated data sequences, including: 1) VINS-Fusion (monocular), 2) VINS-Fusion (stereo), 3) VINS-Fusion (monocular) with ground features, 4) OpenVINS (monocular), 5) ORB-SLAM3 (monocular, with IMU) and 6) the proposed Ground-VIO. For VINS-Fusion (monocular) with ground features, the ground feature processing module is employed, in which ground-truth C-G parameters are applied for comparison. For Ground-VIO, the C-G parameters are unknown and would be estimated online.

For VINS-Fusion-based solutions and Ground-VIO, a 50-ms maximum optimization time limit (single thread, Intel i7-6700K) is set to guarantee real-time processing and provide a more equitable comparison. The maximum feature number of front-end common feature processing is 250, while the maximum number of ground features is set to 40. For an ideal analysis, the semantic images are used to determine the ground region in the simulation tests.

\begin{figure}[!t]
\centering
\includegraphics[width=8.4cm]{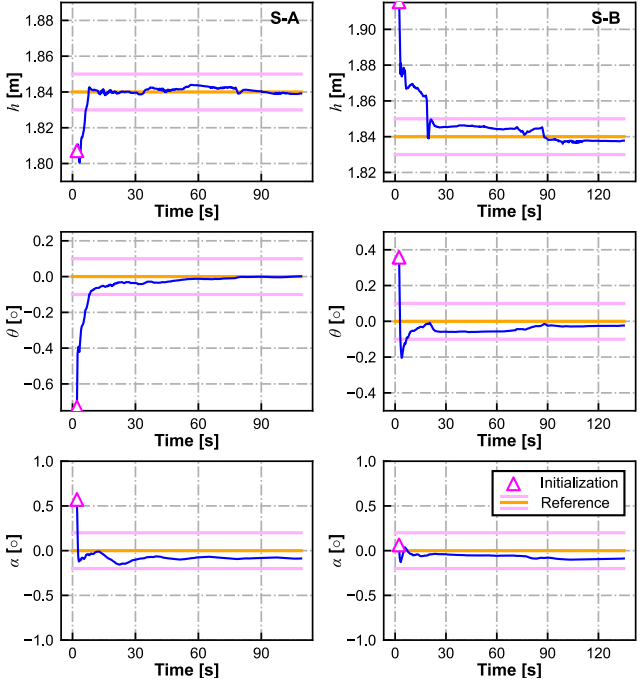}
\caption{Convergence of the C-G parameters during online estimation (left: Seq. S-A, right: Seq. S-B). }
\label{simu_para}
\end{figure}

Firstly. the focus is put on the estimation of the C-G parameters. The convergence of the C-G parameters on Seq. S-A and Seq. S-B is shown in Fig. $\ref{simu_para}$. In the two sequences, the initialization of the C-G parameters could be completed within 10 seconds with $<$0.1 m error of $h$ and $<$1$^\circ$ error of $\theta$ and $\alpha$, as long as enough geometric information is derived by the VIO system. After the initialization, the camera-ground geometric constraints are enabled in the factor graph, and the C-G parameters go on to be refined. It could be found that, with moderate vehicle dynamics and ideal planar grounds in the simulation, the C-G parameters could get converged in a very short time (10 secs for Seq. S-A and 20 secs for Seq. S-B) and achieve good accuracy (0.01 m for $h$, 0.1$^\circ$ for $\theta$ and $\alpha$). Relatively speaking, the convergence performance for Seq. S-A is better, which could be attributed to more available ground features. It is noted that the estimation accuracy of $\alpha$ (roll) is worse than $\theta$ (pitch), which is reasonable considering the region of interest (15 m far and ${\pm}$ 3 m wide) and the fact that only the pitch vibration is compensated (Sect. IV-E). 
%Actually, the accuracy of $\alpha$ is less required to achieve good geometric acccuracy of IPM [], and a $0.2^\circ$ accuracy is enough in most applications.

\begin{figure}[!t]
\centering
\includegraphics[width=8.4cm]{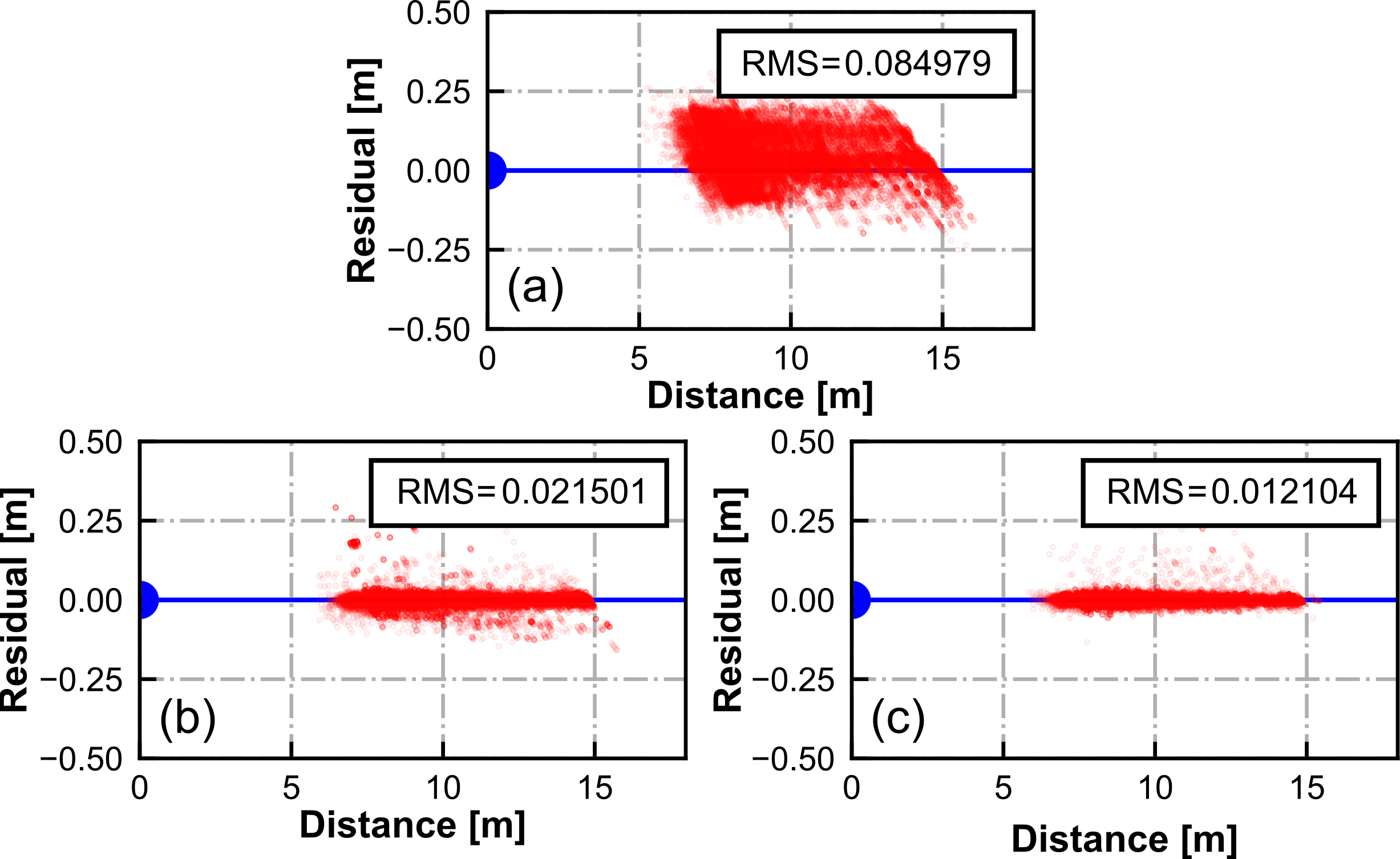}
\caption{Residuals of camera-ground constraints (Eq. (\ref{equation_single_frame})) applying ground-truth C-G parameters in Seq. S-A. The residuals are calculated under different VIO schemes: (a) VINS-Fusion (Mono) without camera-ground constraints. (b) Ground-VIO without any compensation. (c) Ground-VIO with IMU pitch compensation. The X-axis indicates the perception distance of landmarks in the anchor frame. The root mean square (RMS) values are attached besides.}
\label{residuals}
\end{figure}
Secondly, we check the consistency between the landmark depths estimated by VIO and the ground-truth C-G parameters to analyze the model accuracy. To be specific, the residuals of single-frame camera-ground geometric constraints (\ref{equation_single_frame}) are calculated using ground-truth C-G parameters and estimated landmark depths. As shown in Fig. \ref{residuals}, if the constraints are not applied (VINS-Fusion), the estimated landmark depths don't fit the camera-ground geometry well. The distribution of the residuals reflect the error of scale estimation, which could be over the level of $0.1/h$ $\approx$ 5\%. Once the camera-ground geometric constraints are taken into account (Ground-VIO), the residuals are consequently kept to around 0, which indicate an unbiased estimation of the scale. Furthermore, with the local compensation of the vehicle pitch, the noise level of the residuals is significantly lowered. This indicates better accuracy of the compensated model, as it compensates much of the model errors caused by  vehicle dynamics.

\begin{figure}[!t]
\centering
\includegraphics[width=8.4cm]{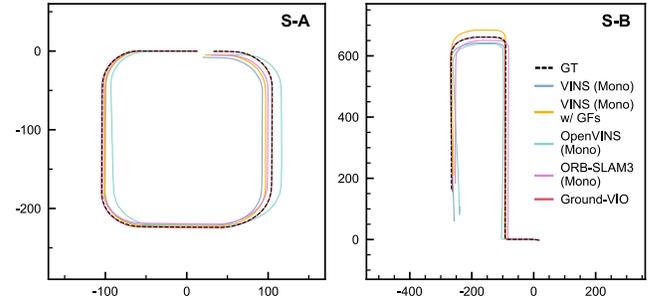}
\caption{Estimated vehicle trajectories of different VIO solutions in the simulation tests (left: Seq. S-A, right: Seq. S-B).}
\label{simu_bev}
\end{figure}
\begin{figure}[!t]
\centering
\includegraphics[width=8.4cm]{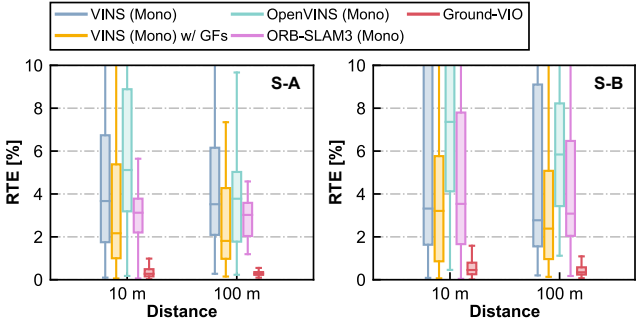}
\caption{Short-term relative translation errors (calculated with 10/100 m intervals) of different VIO solutions in the simulation tests (left: Seq. S-A, right: Seq. S-B).}
\label{simu_box}
\end{figure}

Finally, the pose estimation accuracy of different solutions are investigated. The estimated vehicle trajectories are shown in Fig. \ref{simu_bev}, and the distributions of the relative translation errors are shown in Fig. \ref{simu_box}. Considering that different VIO solutions need different time (several seconds) to initialize the system, we align the estimated vehicle poses at $t$ = 10 s when plotting the trajectories in Fig. \ref{simu_bev}.
It could be found that, the attitude estimations of different solutions are comparable, yet almost all the monocular VIO solutions without C-G constraints show significant translation errors. As the scale observability is closely related to the dynamics in a monocular VIO, the long-time straight motions lead to inevitable scale drifting. The solution of VINS-Fusion (monocular) with ground features slightly improves the translation accuracy by introducing more stable features, but still suffers significant drift of the scale. For Ground-VIO, after the C-G parameters get converged in the beginning dynamic period (with acceleration and rotation), the camera-ground geometry could then provide unbiased information of the metric scale and helps VIO maintain accurate translation estimation. Consequently, the monocular Ground-VIO achieves superior translation estimation performance (relative error $<$ 0.5\%) without introducing any other sensors, which is incredible considering the insufficient dynamics of a ground vehicle in the road environment. 

The statistics of the navigation performance of different VIO solutions are listed in TABLE \ref{table_simu_pos}. The relative translation and rotation errors are calculated by averaging all possible subsequences of length (100, ..., 800) meters, referring to\cite{ref_kitti}. The absolute trajectory error is calculated referring to\cite{ref_gc2}.

\begin{table}[h]
\caption{Pose Estimation Errors In the Simulation Tests (S-A and S-B, Online Calibration).}
\label{table_simu_pos}
\begin{tabular}{l|lll|lll}
\bottomrule
\multirow{2}{*}{\makecell[c]{Method}} & \multicolumn{3}{c|}{Seq. S-A} & \multicolumn{3}{c}{Seq. S-B}            \\ \cline{2-7} 
&\makecell[c]{ $t_{rel}$ \\ ($\%$)} & \makecell[c]{$r_{rel}$\textsuperscript{1}} &\makecell[c]{ $t_{abs}$ \\ ($m$)} &\makecell[c]{ $t_{rel}$ \\ ($\%$)} & \makecell[c]{$r_{rel}$\textsuperscript{1}} &\makecell[c]{ $t_{abs}$ \\ ($m$)}   \\
\hline
VINS-Fusion (Mono)                       &  \makecell[c]{3.81}    &  \makecell[c]{0.14}  &  \makecell[c]{7.78}   &  \makecell[c]{5.05}  & \makecell[c]{0.21}  & \makecell[c]{22.25} \\

\makecell[l]{VINS-Fusion (Mono) \\  \textbackslash w GFs}
&  \makecell[c]{2.60}
&  \makecell[c]{0.13}
&  \makecell[c]{5.58}
&  \makecell[c]{2.98}
&  \makecell[c]{0.15}
&  \makecell[c]{12.34}
\\
OpenVINS (Mono)                          &  \makecell[c]{2.73}    &  \makecell[c]{0.68}  &  \makecell[c]{4.22}   &  \makecell[c]{5.30}  & \makecell[c]{0.71}  & \makecell[c]{19.45}  \\
ORB-SLAM3 (Mono)                         &  \makecell[c]{2.27}    &  \makecell[c]{0.11}  &  \makecell[c]{3.88}   &  \makecell[c]{3.09}  & \makecell[c]{0.14}  & \makecell[c]{9.40}  \\
Ground-VIO                               &  \makecell[c]{\textbf{0.24}}  &  \makecell[c]{\textbf{0.10}}  &  \makecell[c]{\textbf{0.36}}  &    \makecell[c]{\textbf{0.38}}& \makecell[c]{\textbf{0.11}}  & \makecell[c]{\textbf{1.43}} \\ 
\toprule
\end{tabular}
\begin{tablenotes}
\item[1] Unit: $^\circ/100\ m$.
\end{tablenotes}
\end{table}

\begin{figure}[!t]
\centering
\includegraphics[width=5.4cm]{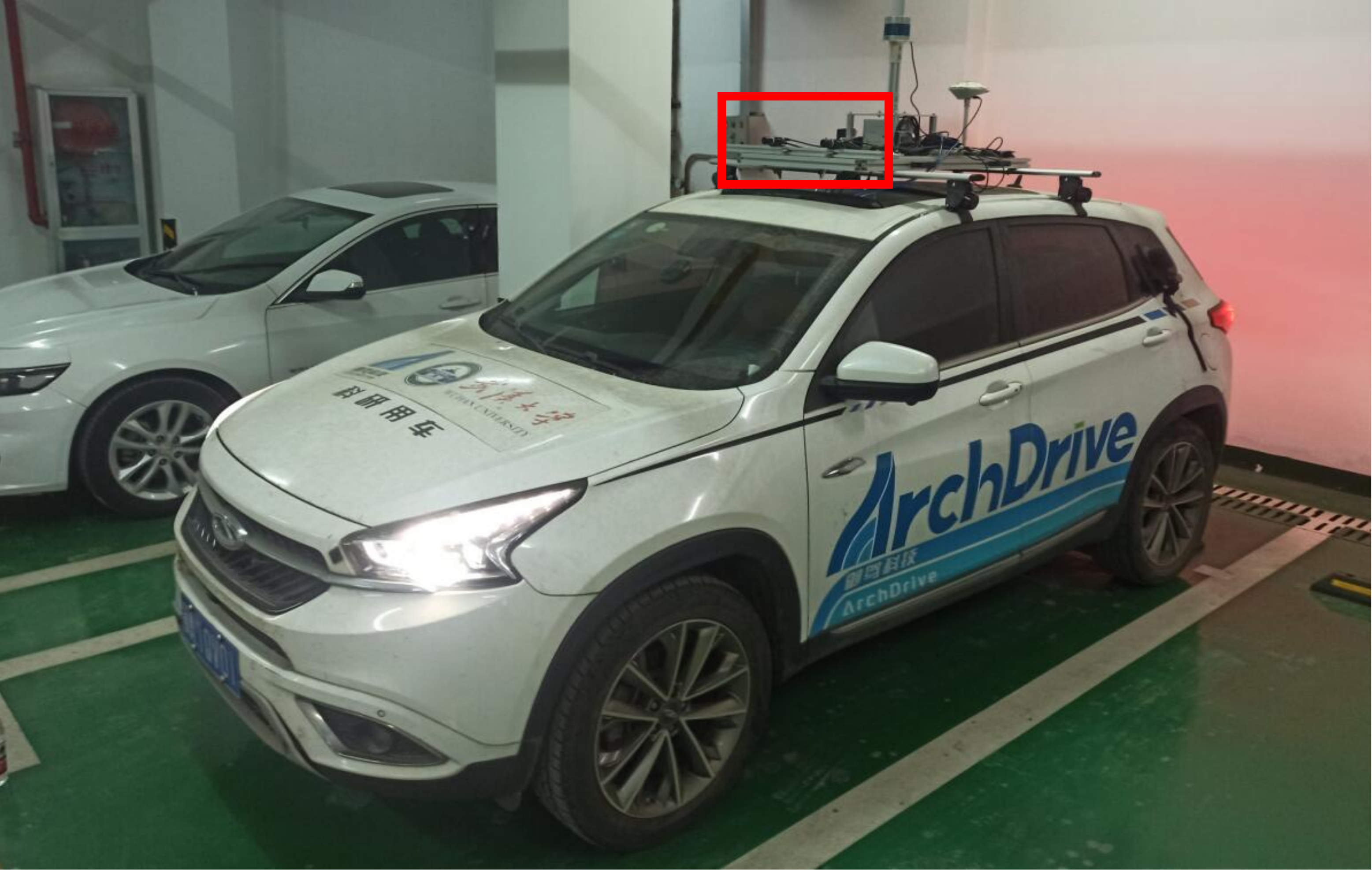}
\caption{Appearance of the experimental vehicle.}
\label{appearance}
\end{figure} 

\section{Real-World Experiments}
\begin{figure*}[!t]
\centering
\includegraphics[width=17.2cm]{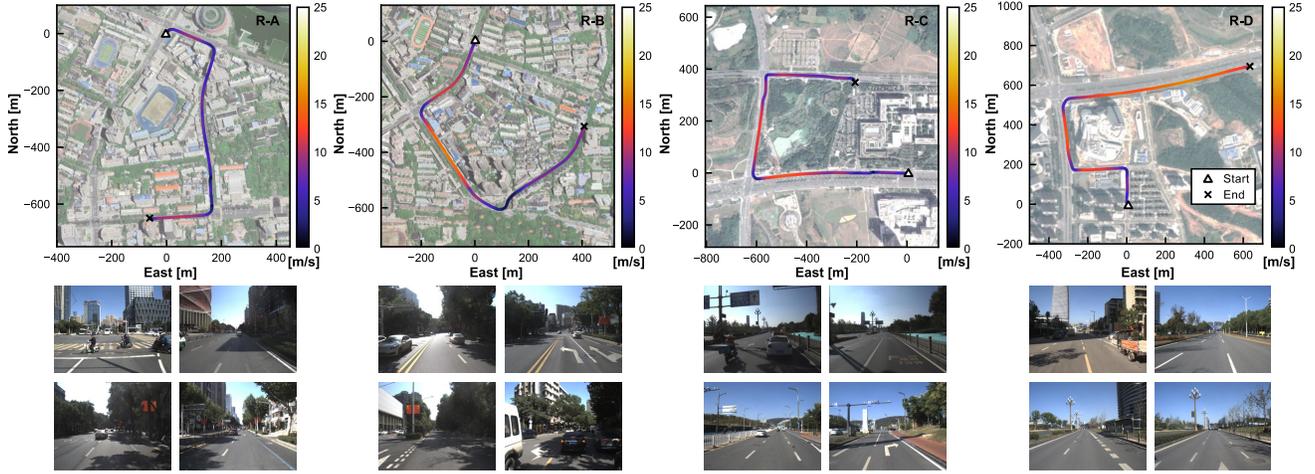}
\caption{Vehicle trajectories and example images in Seq. R-A, R-B, R-C and R-D. The colorbar indicates the vehicle speed (m/s). Seq. R-A and Seq. R-B are under the urban scenario, with narrow roads, abundant environmental textures and relatively low speed. Seq. R-C and Seq. R-B are mostly on the highway, with broad roads, fewer buildings and moderate speed. }
\label{expr}
\end{figure*}

Real-world experiments were conducted on Oct. 12, 2022 to evaluate the performance of the proposed system under typical vehicular scenarios, including urban roads and highways. The appearance of the experimental vehicle is shown in Fig. \ref{appearance}. The experimental platform is equipped with two Flir BFS-PGE-31S4C cameras, a low-cost ADIS16470 MEMS IMU, a tactical grade XW-GI7660 IMU and a Septentrio AsteRx4 GNSS receiver. The data from the tactical grade IMU and the GNSS receiver (with base station availability) are post-processed to generate the reference trajectory. The specifications of the used IMUs are listed in TABLE \ref{table_imu}.

\begin{table}[h]
\caption{Specifications of the Used IMUs.}
\label{table_imu}
\centering
\begin{tabular}{l|ll|ll}
\hline
\multirow{2}{*}{\makecell[c]{IMU}} & \multicolumn{2}{c|}{Noise Density} & \multicolumn{2}{c}{Bias Stability}            \\ \cline{2-5} 
& \makecell[c]{Gyro.\\($^\circ/\sqrt{hr}$)}  & \makecell[c]{Accel.\\($m/s/\sqrt{hr}$)} 
& \makecell[c]{Gyro.\\($^\circ/hr$)} &\makecell[c]{Accel.\\($mGal$)}   \\
\hline
ADIS16470&\makecell[c]{0.34} & \makecell[c]{0.037}  & \makecell[c]{8}  & \makecell[c]{1300}  \\
XW-GI7660&  \makecell[c]{-}    &  \makecell[c]{-}  &  \makecell[c]{0.3}   &  \makecell[c]{100}    \\ 
\hline
\end{tabular}
\end{table}

The reason to use self-collected datasets is for better representativeness of the evaluation, i. e., using low-cost visual-inertial sensor scheme under realistic vehicular scenarios, and especially focusing on feature-lacking highway scenarios with limited dynamics. For the visibility of our work, we will make the experimental data public available.

Notice that in the real-world experiments,  we don't apply a semantic segmentation module but rely on the system itself to distinguish the ground features and resist possible outliers.

\begin{figure*}[!t]
\centering
\includegraphics[width=17.0cm]{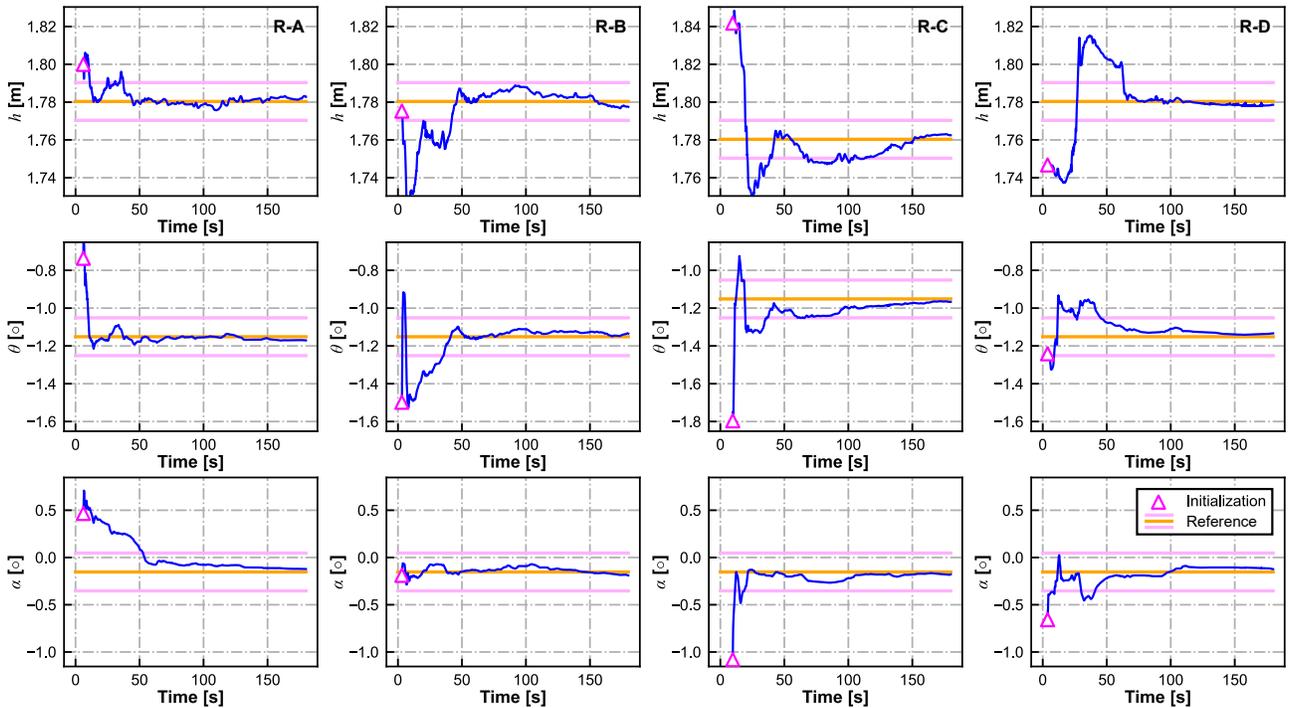}
\caption{Convergence of the C-G parameters during online estimation in Seq. R-A, R-B, R-C and R-D.}
\label{expr_para}
\end{figure*} 

\subsection{VIO with Unknown C-G Parameters}

In this part, we evaluate the proposed system under the condition that the C-G parameters are completely unknown. In this case, the C-G parameters would be initialized online and continuously estimated during the data periods. Four 180-sec data sequences with moderate vehicle dynamics, namely Seq. R-A, R-B, R-C and R-D, are used for the evaluation, as shown in Fig. \ref{expr}. Different solutions of VIO are tested on the data sequences. Compared to the simulation test, state-of-art VIO implementations with stereo camera setups are considered in this part to investigate the best achievable VIO performance in these real-world road environments.

The convergence of the C-G parameters is shown in Fig. \ref{expr_para}. For the four sequences, the final estimation results of the C-G parameters show good consistency. Later in this part, the average value of the four sets of obtained C-G parameters is taken as the reference, which is (1.7803  m,\ -1.151$^\circ$,\ -0.153$^\circ$) for  ($h$, $\theta$, $\alpha$). It is found from Fig. \ref{expr_para} that, the initialization of the parameters could be finished in a few seconds, and the initial accuracy is similar to the simulation tests (0.1 m for $h$, 1$^\circ$ for $\theta$, $\alpha$). Yet differently, the convergence of the C-G parameters is slower than the simulation tests. To be specific, around 30$\sim$60 seconds  are needed to obtain ideal accuracy of the C-G parameters (0.01 m, 0.1$^\circ$, 0.2$^\circ$ for $h$, $\theta$, $\alpha$). This could be attributed to more complex road conditions and smaller IMU excitation in the real-world experiments, which affect both the camera-ground geometric constraint and the monocular VIO itself. Roughly speaking, better observability of the VIO system, sufficient ground features and smooth road surface could contribute to faster convergence of the C-G parameters.

\begin{figure}[!t]
\centering
\includegraphics[width=8.4cm]{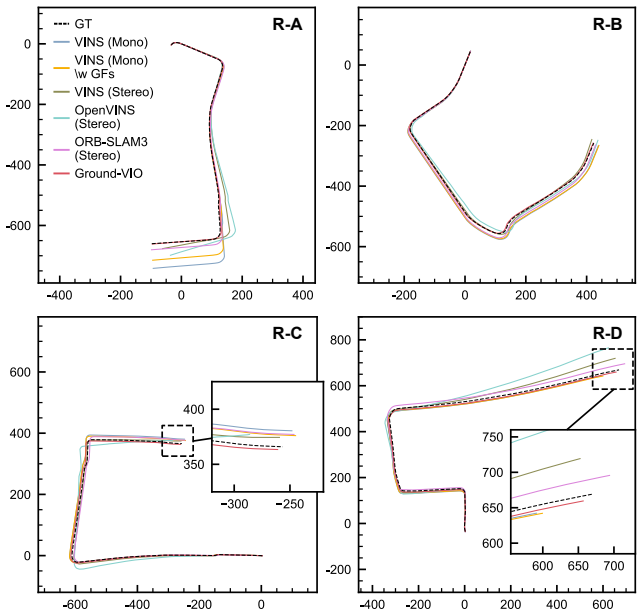}
\caption{Estimated vehicle trajectories of different VIO solutions in the simulation tests (left: Seq. R-A, right: Seq. R-B).}
\label{expr_bev}
\end{figure}

\begin{figure}[!t]
\centering
\includegraphics[width=8.4cm]{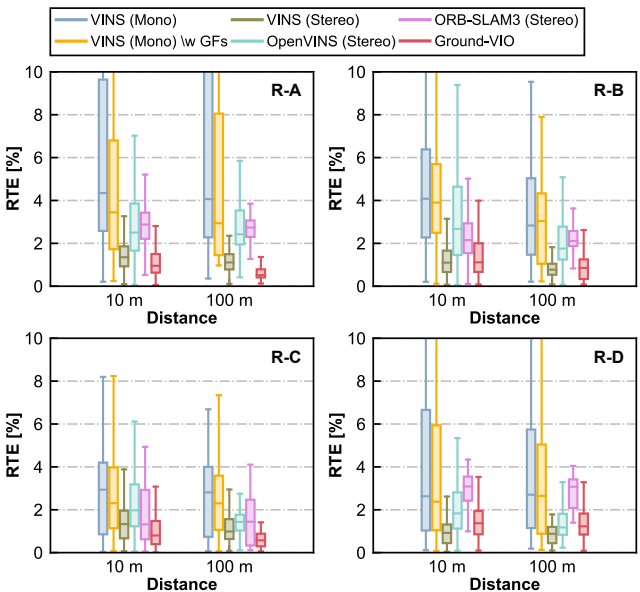}
\caption{Short-term relative translation errors (calculated with 10/100 m intervals) of different VIO solutions in the simulation tests (left: Seq. R-A, right: Seq. R-B).}
\label{expr_box}
\end{figure}

The focus is then put on the pose estimation performance. Fig. \ref{expr_bev} and Fig. \ref{expr_box} show the estimated vehicle trajectories and relative translation errors of different VIO schemes. The detailed statistics of the VIO performance are listed in TABLE \ref{table_pos_abcd}. Similar to the simulation tests, the monocular VIOs (except Ground-VIO) obtain good attitude estimation, but show bad performance on the translation error due to the drift of the scale, which is significant during long-time straight motions. 
Comparatively, the stereo VIOs perform much better on the relative translation errors, but they still undergo significant pose errors as could be seen in Fig. \ref{expr_bev}. To be specific, the filter-based OpenVINS (stereo) undergoes relatively large heading errors on the four sequences, while the optimization-based VINS-Fusion (stereo) performs bad on Seq. R-A and R-D. The phenomenon that monocular VIO could outperform stereo VIO on attitude estimation could also be found in\cite{ref_vinsfusion}. The ORB-SLAM3 (stereo, with IMU) scheme, although maintaining good heading estimation, undergoes non-negligible position drifting. The good attitude estimation performance could be attributed to the map-centered design of ORB-SLAM3, whose superiority is verified in\cite{ref_orbslam3}. However, the road environment, with insufficient stable features and moving objects, has caused difficulty for it to achieve ideal translation estimation.

\begin{table*}[h]
\caption{Pose Estimation Performance on R-A, R-B, R-C and R-D Sequences (Online Calibration).}
\label{table_pos_abcd}
\begin{tabular}{l|lll|lll|lll|lll}
\hline
\multirow{2}{*}{\makecell[c]{Method}} & \multicolumn{3}{c|}{{Seq. R-A}} & \multicolumn{3}{c|}{Seq. R-B}  & \multicolumn{3}{c|}{{Seq. R-C}} & \multicolumn{3}{c}{Seq. R-D}            \\ \cline{2-13} 
&\makecell[c]{ $t_{rel}$ \\ ($\%$)} & \makecell[c]{$r_{rel}$\\ ($^\circ/100\ m$)} &\makecell[c]{ $t_{abs}$ \\($m$)} 
&\makecell[c]{ $t_{rel}$ \\ ($\%$)} & \makecell[c]{$r_{rel}$\\ ($^\circ/100\ m$)} &\makecell[c]{ $t_{abs}$ \\ ($m$)} 
&\makecell[c]{ $t_{rel}$ \\ ($\%$)} & \makecell[c]{$r_{rel}$\\ ($^\circ/100\ m$)} &\makecell[c]{ $t_{abs}$ \\ ($m$)} 
&\makecell[c]{ $t_{rel}$ \\ ($\%$)} & \makecell[c]{$r_{rel}$\\ ($^\circ/100\ m$)} &\makecell[c]{ $t_{abs}$ \\ ($m$)}    \\
\hline
VINS-Fusion (Mono)
&\makecell[c]{10.4}    &  \makecell[c]{0.17}  &  \makecell[c]{33.9}   
&\makecell[c]{3.06}    &  \makecell[c]{0.13}  &  \makecell[c]{9.15}
&  \makecell[c]{2.17}  & \makecell[c]{\textbf{0.08}}   & \makecell[c]{7.12} 
&\makecell[c]{2.96}    & \makecell[c]{0.09}   & \makecell[c]{18.0}\\
\makecell[l]{VINS-Fusion (Mono) \\ \textbackslash w GFs}
&\makecell[c]{6.84}    &  \makecell[c]{0.16}  &  \makecell[c]{22.3}
&\makecell[c]{2.99}    &  \makecell[c]{\textbf{0.11}}  &  \makecell[c]{8.81}  
&  \makecell[c]{2.09}  & \makecell[c]{0.10}   & \makecell[c]{6.29}   
&  \makecell[c]{2.62}  & \makecell[c]{0.09}   & \makecell[c]{16.1} \\
VINS-Fusion (Stereo)                      
&\makecell[c]{2.03}    &  \makecell[c]{0.80}  &  \makecell[c]{4.44}
&\makecell[c]{1.15}    &  \makecell[c]{0.40}  &  \makecell[c]{3.74}   
&  \makecell[c]{0.88}  & \makecell[c]{0.19}   & \makecell[c]{3.22}   
&  \makecell[c]{\textbf{1.08}}  & \makecell[c]{0.26}   & \makecell[c]{8.01} \\
OpenVINS (Stereo)                         
&\makecell[c]{5.11}    &  \makecell[c]{2.07}  &  \makecell[c]{10.6}
&\makecell[c]{2.37}    &  \makecell[c]{0.69}  &  \makecell[c]{3.92}   
&  \makecell[c]{2.11}  & \makecell[c]{0.65}   & \makecell[c]{7.11}   
&  \makecell[c]{2.05}  & \makecell[c]{0.72}   & \makecell[c]{12.7}  \\
ORB-SLAM3 (Stereo)                        
&\makecell[c]{2.45}    &  \makecell[c]{0.22}  &  \makecell[c]{6.74}
&\makecell[c]{2.19}    &  \makecell[c]{0.24}  &  \makecell[c]{5.80}   
&  \makecell[c]{1.36}  & \makecell[c]{0.15}   & \makecell[c]{4.52}   
&  \makecell[c]{2.58}  & \makecell[c]{0.10}   & \makecell[c]{12.0}  \\
Ground-VIO                               
&\makecell[c]{\textbf{0.42}}     &  \makecell[c]{\textbf{0.14}}  &  \makecell[c]{\textbf{0.67}}
&\makecell[c]{\textbf{0.72}}              &  \makecell[c]{\textbf{0.11}}           &  \makecell[c]{\textbf{1.30}}   
&  \makecell[c]{\textbf{0.48}}            & \makecell[c]{0.09}            & \makecell[c]{\textbf{1.28}}  
&    \makecell[c]{\textbf{1.08}} & \makecell[c]{\textbf{0.08}}   & \makecell[c]{\textbf{5.15}} \\
\hline
\end{tabular}
\begin{tablenotes}
\end{tablenotes}
\end{table*}

In constrast, the proposed Ground-VIO shows good translation estimation performance with the help of the camera-ground geometric constraints. Although the C-G parameters are unknown at the beginning, the vehicle dynamics are able to make them converge and continuously take effect in the remaining period. It is verified that, the camera-ground geometry, like stereo vision but in a different way, could help maintain precise and unbiased scale estimation in realistic vehicular scenarios. Generally, the Ground-VIO could achieve comparable relative translation error (0.5\%$\sim$1.0\%) with state-of-art stereo VIO schemes, and the attitude estimation performance is even better. Thus, the Ground-VIO achieves the smallest position drift on almost all the four sequences. 

In all, with moderate vehicle dynamics, the proposed Ground-VIO is able to online calibrate the C-G parameters and obtain good pose estimation accuracy simultaneously.

\subsection{Pre-Calibrated VIO under Challenging Scenarios}
It has been mentioned that, the online calibration of the C-G parameters relys on the vehicle dynamics, since it needs the observability of VIO to extract metric-scale environmental structure. Fortunately, with pre-calibration of the C-G parameters, the VIO performance could also be greatly improved even under dynamic-insufficient scenarios. Actually, the pre-calibration is not hard, since it could be automatically finished when moderate vehicle dynamics are available, as verified in Sect. VI-A.

In this section, two highway data sequences, namely Seq. R-E and Seq. R-F, are used to test the system performance with pre-calibrated C-G parameters. The vehicle trajectories and the representative images are shown in Fig. \ref{expr}. These two data sequences are extremely challenging for VIO, with limited dynamics, insufficient environmental features and high vehicle speed. These conditions could cause difficulty in both feature tracking and the observability of the VIO system. For Ground-VIO, the pre-calibrated C-G parameters are obtained from the online estimation results in Sect. VI-A.

The estimated vehicle trajectories and relative translation error distributions are shown in Fig. \ref{exprgo2_bev} and Fig. \ref{exprgo_bev}. Despite our best efforts, some schemes can't work properly on the two sequences. To be specific, the OpenVINS (stereo) scheme can't successfully initialize on both sequences and the ORB-SLAM3 (stereo) scheme fails on Seq. R-F because of the difficulty in ORB feature matching, as the environmental textures are either weak or highly repetitive (e.g. building windows, guardrails).

The pose estimation results are presented in Fig. \ref{exprgo2_bev} and Fig. \ref{exprgo_bev}. As shown in Fig. \ref{exprgo2_bev}, the monocular VIOs perform bad on the translation estimation on Seq. R-E, reaching a relative error over 10\%. It is unexpected that state-of-art stereo VIO schemes are also unable to achieve good pose estimation on the sequences, despite the fact that the stereo vision could provide accurate scale  information in principle. This could be mainly due to the lack of high-quality visual features in the highway environment, and the stereo matching even increases the risk of introducing gross errors. In contrast, with the pre-calibrated C-G parameters, the proposed Ground-VIO achieves an incredible 1\% relative translation error (average). The camera-ground geometry not only provides unbiased scale information to the monocular VIO system, but also provides stable features via the specially designed ground feature processing module. In the highway environment, most visual features are faraway which can't provide accurate translation information. The ground feature processing module makes it possible to fully utilize the landmarks on the road (e. g. road markings, shadows) to mitigate the ill-conditioning, and what makes sense is that these landmarks depths are almost known. As a result, the estimation of vehicle velocity is effectively constrained and contributes to accurate pose estimation.

\begin{figure}[!t]
\centering
\includegraphics[width=8.4cm]{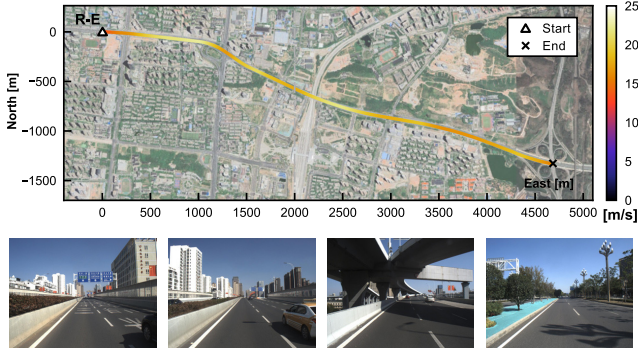}
\caption{Vehicle trajectories and example images in Seq. R-E. The data sequence is on the highway with a relatively high speed. Nearby buildings, trees and traffic signs provide a limited number of visual features.}
\label{fig_1}
\end{figure} 
\begin{figure}[!t]
\centering
\includegraphics[width=8.4cm]{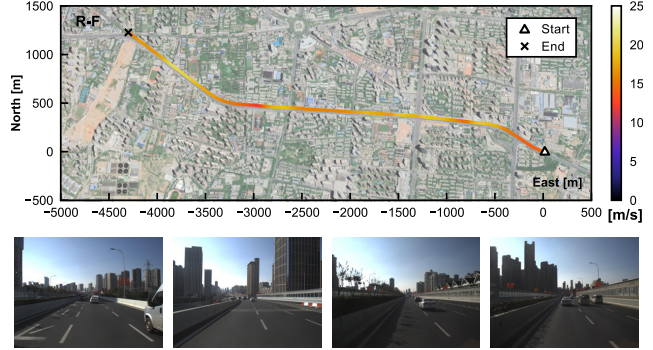}
\caption{Vehicle trajectories and example images in Seq. R-F. The data sequence is on the highway with moderate/high speed. The buildings in the view are faraway and the textures of the surrounding objects (buildings, guardrails and bushes) are mostly repetitive.}
\label{fig_1}
\end{figure} 
\begin{figure}[!t]
\centering
\includegraphics[width=8.4cm]{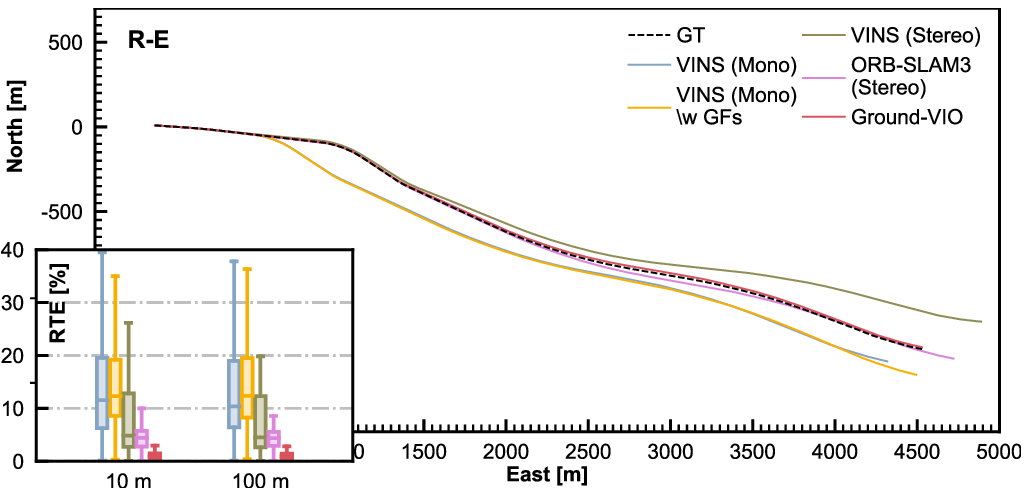}
\caption{Estimated vehicle trajectories and relative translation errors of different VIO solutions on Seq. R-E.}
\label{exprgo2_bev}
\end{figure} 
\begin{figure}[!t]
\centering
\includegraphics[width=8.4cm]{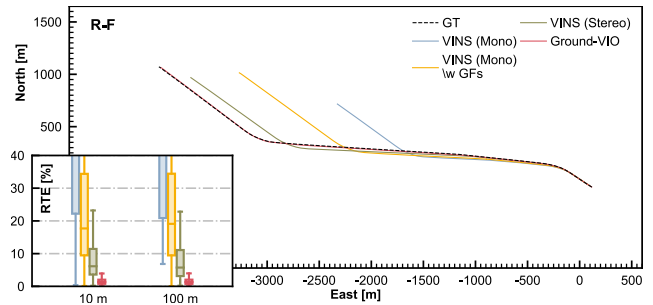}
\caption{Estimated vehicle trajectories and relative translation errors of different VIO solutions on Seq. R-F.}
\label{exprgo_bev}
\end{figure} 

Similar results could be found in the more challenging Seq. R-F. As shown in Fig. \ref{exprgo_bev}, the Ground-VIO greatly outperforms state-of-art monocular and stereo VIO schemes, achieving 1\% relative translation error (average).

The statistics of the pose estimation error on Seq. R-E and R-F are listed in TABLE \ref{table_exprgo}.

\begin{table}[h]
\caption{Pose Estimation Performance on R-E and R-F Data Sequences (Pre-calibrated).}
\label{table_pos}
\begin{tabular}{l|lll|lll}
\hline
\multirow{2}{*}{\makecell[c]{Method}} & \multicolumn{3}{c|}{Seq. S-A} & \multicolumn{3}{c}{Seq. S-B}            \\ \cline{2-7} 
&\makecell[c]{ $t_{rel}$ \\ ($\%$)} & \makecell[c]{$r_{rel}$\textsuperscript{1}} &\makecell[c]{ $t_{abs}$ \\ ($m$)} &\makecell[c]{ $t_{rel}$ \\ ($\%$)} & \makecell[c]{$r_{rel}$\textsuperscript{1}} &\makecell[c]{ $t_{abs}$ \\ ($m$)}   \\
\hline
VINS-Fusion (Mono)                        &\makecell[c]{15.3} & \makecell[c]{0.12}  & \makecell[c]{*}  & \makecell[c]{41.7} & \makecell[c]{0.16}  & \makecell[c]{*} \\
\makecell[l]{VINS-Fusion (Mono) \\ \textbackslash w GFs}&\makecell[c]{17.2} & \makecell[c]{0.12}  & \makecell[c]{*}  & \makecell[c]{22.4} & \makecell[c]{0.11}  & \makecell[c]{*} \\
VINS-Fusion (Stereo)                     &  \makecell[c]{7.2}    &  \makecell[c]{0.22}  &  \makecell[c]{*}   &  \makecell[c]{8.27}   & \makecell[c]{0.15}  & \makecell[c]{*} \\
ORB-SLAM3 (Stereo)                      &  \makecell[c]{4.25}    &  \makecell[c]{0.13}  &  \makecell[c]{67.6}   &  \makecell[c]{-}   & \makecell[c]{-}  & \makecell[c]{-} \\
Ground-VIO                             &  \makecell[c]{\textbf{0.85}}  &  \makecell[c]{\textbf{0.05}}  &  \makecell[c]{\textbf{6.96}}  &    \makecell[c]{\textbf{1.21}}& \makecell[c]{\textbf{0.06}}  & \makecell[c]{\textbf{5.72}} \\ 
\hline
\end{tabular}
\begin{tablenotes}
\item[1] Unit: $^\circ/100\ m$. \item[*] Greater than $ 100\ m$.
\end{tablenotes}
\label{table_exprgo}
\end{table}

\subsection{IPM Calibration Performance}
To some extent, the online estimation of C-G parameters is equivalent to online calibrating IPM, which is widely used in vehicle perception. 
In this part, apart from the odometry performance, the effectiveness of the online IPM calibration is investigated qualitatively. 

To be specific, the estimated C-G parameters in Sect. VI-A are used for IPM processing of an image sequence. Through IPM, the images are transformed into metric-scale point clouds with color information. Based on the camera poses obtained by Ground-VIO, the point clouds could be merged together in a reference frame. The consistency of the merged point could directly reflects the accuracy of IPM.

\begin{figure}[!t]
\centering
\includegraphics[width=8.4cm]{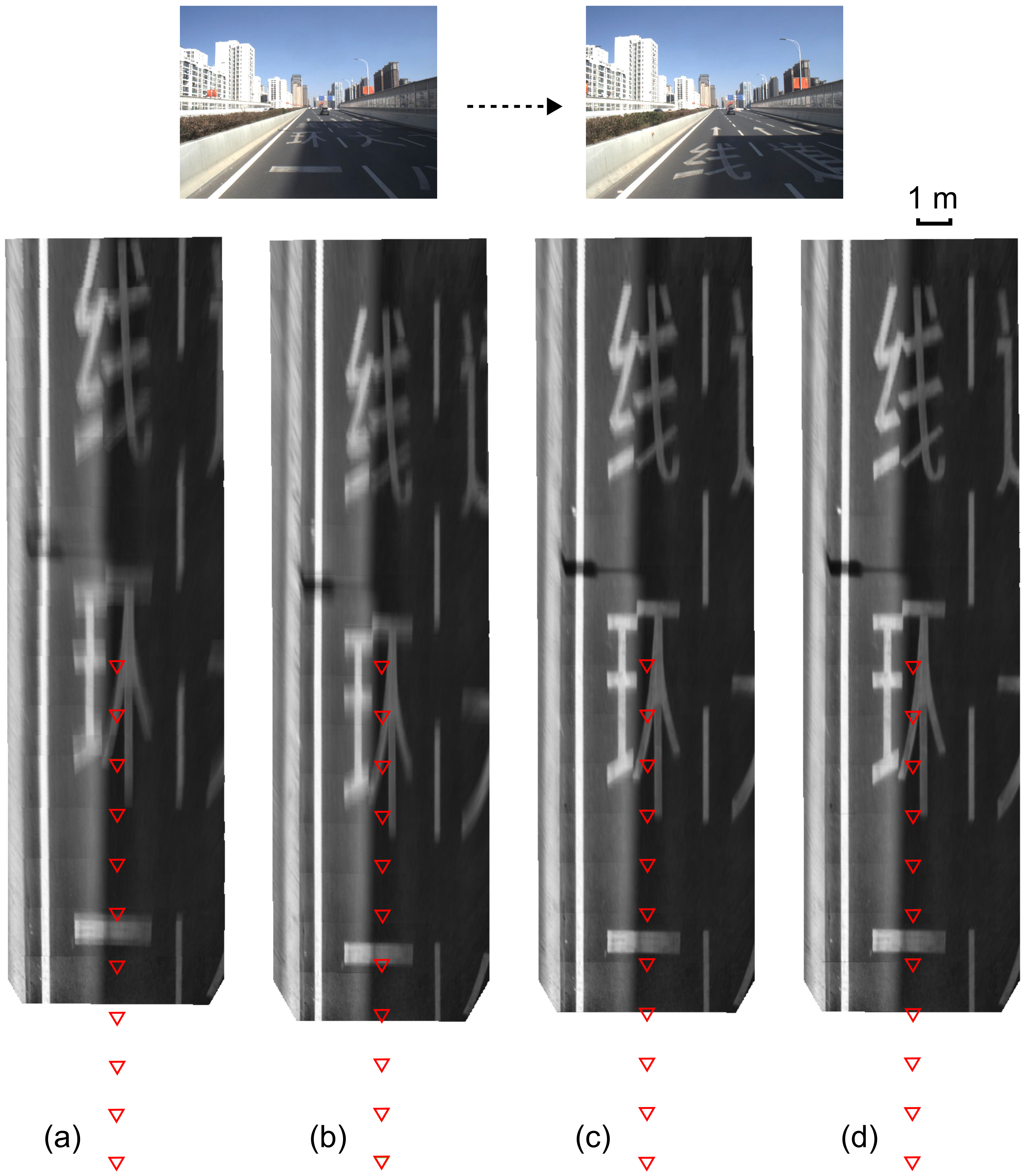}
\caption{Merged IPM point clouds obtained from 11 successive images, based on different C-G parameters: (a) With 1$^{\circ}$ error of $\theta$. (b) With 0.1 m error of $h$. (c) Calibrated C-G parameters. (d) Calibrated C-G parameters with IMU pitch compensation. The red triangles indicate the camera poses.}
\label{IPM}
\end{figure}

Fig. \ref{IPM} shows the merged point clouds based on different C-G parameters. It could be seen that, with residual error on either attitude or height component of the C-G parameters, the merged point cloud is fuzzy and has stitching errors. This reflects the inconsistency of multiple point clouds, which further indicates that the generated IPM point clouds are not geometrically accurate. Roughly speaking, the longitudinal errors could reach over meter-level. In contrast, with calibrated C-G parameters, the merged point cloud is much more consistent and less fuzzy. Furthermore, once the IMU pitch compensation is applied, the details of the point cloud become clearer, which verifies the accuracy of the camera-ground geometric model. After the IPM calibration, a 10$\sim$15 meter effective perception range, with decimeter to centimeter level accuracy, could be expected.

In all, the proposed algorithm provides an approach to online calibrate the IPM parameters of vehicle-mounted cameras, which is based on only monocular visual-inertial data without the need of extra infrastructure. 

%
%\subsection{Additional Tests on Public Datasets}
%For the representativeness of the evaluation, we have 
%XX- XX- XX- 
\section{Conclusion}
In this work, we presented Ground-VIO, which introduces the camera-ground geometry into monocular VIO to improve the odometry performance. The proposed works well with either unknown or pre-calibrated C-G parameters, achieving comparable or even better odometry accuracy than state-of-art stereo VIOs in vehicular scenarios. The method is expected to significantly improve the practicability of VIO applied in intelligent vehicle applications, which could work as an effective supplement to existing vehicle navigation schemes.

Besides, the proposed method provides an efficient way for online IPM calibration based on only monocular visual-inertial data. The auto-calibration doesn't need extra infrastructure and could handle long-term change of the sensor alignment, which is meaningful for better vehicle perception. It is our future interest to apply this technique in vision-based crowd-sourced mapping applications.

\section*{Acknowledgments}
The implemented Ground-VIO is part of the GREAT (GNSS+ REsearch, Application and Teaching) software developed by the GREAT Group, School of Geodesy and Geomatics, Wuhan University.

\vfill

\end{document}